\pdfoutput=1
\newif\ifarxiv
\arxivtrue
\iftrue\pdfmapfile{+classico.map}\fi
\newif\ifafour
\afourfalse
\newif\iftypodisclaim 
\typodisclaimtrue

\documentclass[\ifafour a4paper,12pt,\else a5paper,10pt,\fi
onecolumn,oneside,article,
british%
]{memoir}
\newcommand*{\firstdraft}{27 May 2022}
\newcommand*{\firstpublished}{\firstdraft}
\newcommand*{\updated}{\ifarxiv 21 February 2023\else\today\fi}
\newcommand*{\propertitle}{Does the evaluation stand up to evaluation?\\ {\Large A first-principle approach to the evaluation of classifiers}}
\newcommand*{\pdftitle}{\propertitle}
\newcommand*{\headtitle}{Does the evaluation stand up to evaluation?}
\newcommand*{\pdfauthor}{K. Dyrland, A. S. Lundervold, P.G.L.  Porta Mana}
\newcommand*{\headauthor}{Dyrland, Lundervold, Porta Mana}
\newcommand*{\reporthead}{\ifarxiv\else Open Science Framework \href{https://doi.org/10.31219/osf.io/7rz8t}{\textsc{doi}:10.31219/osf.io/7rz8t}\fi}

\usepackage[T1]{fontenc} 
\input{glyphtounicode} \pdfgentounicode=1

\usepackage[utf8]{inputenx}

\usepackage{textcomp}

\usepackage{amsmath}

\usepackage{mathtools}
\usepackage{empheq}
\newcommand*{\widefbox}[1]{\fbox{\hspace{1em}#1\hspace{1em}}}

\usepackage{amssymb}

\usepackage{amsxtra}

\usepackage[main=british]{babel}

\usepackage[autostyle=false,autopunct=false,english=british]{csquotes}
\setquotestyle{british}

\usepackage{amsthm}
\makeatletter
\def\@endtheorem{\endtrivlist}
\makeatother

\theoremstyle{remark}

\newtheoremstyle{innote}{\parsep}{\parsep}{\footnotesize}{}{}{}{0pt}{}
\theoremstyle{innote}

\usepackage[shortlabels,inline]{enumitem}
\SetEnumitemKey{para}{itemindent=\parindent,leftmargin=0pt,listparindent=\parindent,parsep=0pt,itemsep=\topsep}
\setlist{itemsep=0pt,topsep=\parsep}
\setlist[enumerate,2]{label=(\roman*)}
\setlist[enumerate]{label=(\alph*),leftmargin=1.5\parindent}
\setlist[itemize]{leftmargin=1.5\parindent}
\setlist[description]{leftmargin=1.5\parindent}

\usepackage[babel,theoremfont,largesc]{newpxtext}

\usepackage[bigdelims,nosymbolsc
]{newpxmath}
\makeatletter
\def\re@DeclareMathSymbol#1#2#3#4{%
    \let#1=\undefined
    \DeclareMathSymbol{#1}{#2}{#3}{#4}}
\re@DeclareMathSymbol{\bigoplusop}{\mathop}{largesymbols}{"4C}
\re@DeclareMathSymbol{\bigotimesop}{\mathop}{largesymbols}{"4E}
\re@DeclareMathSymbol{\sumop}{\mathop}{largesymbols}{"50}
\re@DeclareMathSymbol{\prodop}{\mathop}{largesymbols}{"51}
\re@DeclareMathSymbol{\bigcupop}{\mathop}{largesymbols}{"53}
\re@DeclareMathSymbol{\bigcapop}{\mathop}{largesymbols}{"54}
\re@DeclareMathSymbol{\bigwedgeop}{\mathop}{largesymbols}{"56}
\re@DeclareMathSymbol{\bigveeop}{\mathop}{largesymbols}{"57}
\re@DeclareMathSymbol{\bigtimesop}{\mathop}{largesymbolsPXA}{"10}
\makeatother
\DeclareFontFamily{U}{egreek}{\skewchar\font'177}%
\DeclareFontShape{U}{egreek}{m}{n}{<-6>s*[1]eurm5 <6-8>s*[1]eurm7 <8->s*[1]eurm10}{}%
\DeclareFontShape{U}{egreek}{m}{it}{<->s*[1]eurmo10}{}%
\DeclareFontShape{U}{egreek}{b}{n}{<-6>s*[1]eurb5 <6-8>s*[1]eurb7 <8->s*[1]eurb10}{}%
\DeclareFontShape{U}{egreek}{b}{it}{<->s*[1]eurbo10}{}%
\DeclareSymbolFont{egreeki}{U}{egreek}{m}{it}%
\SetSymbolFont{egreeki}{bold}{U}{egreek}{b}{it}
\DeclareSymbolFont{egreekr}{U}{egreek}{m}{n}%
\SetSymbolFont{egreekr}{bold}{U}{egreek}{b}{n}
\DeclareFontFamily{U}{egreekx}{\skewchar\font'177}
\DeclareFontShape{U}{egreekx}{m}{n}{%
       <-7.5>s*[0.9]euex7%
    <7.5-8.5>s*[0.9]euex8%
    <8.5-9.5>s*[0.9]euex9%
    <9.5->s*[0.9]euex10%
}{}
\DeclareSymbolFont{egreekx}{U}{egreekx}{m}{n}
\DeclareMathSymbol{\sumop}{\mathop}{egreekx}{"50}
\DeclareMathSymbol{\prodop}{\mathop}{egreekx}{"51}
\DeclareMathSymbol{\coprodop}{\mathop}{egreekx}{"60}
\makeatletter
\def\sum{\DOTSI\sumop\slimits@}
\def\prod{\DOTSI\prodop\slimits@}
\def\coprod{\DOTSI\coprodop\slimits@}
\makeatother
\DeclareMathSymbol{\varpartial}{\mathalpha}{egreeki}{"40}
\DeclareMathSymbol{\partialup}{\mathalpha}{egreekr}{"40}
\DeclareMathSymbol{\alpha}{\mathalpha}{egreeki}{"0B}
\DeclareMathSymbol{\beta}{\mathalpha}{egreeki}{"0C}
\DeclareMathSymbol{\gamma}{\mathalpha}{egreeki}{"0D}
\DeclareMathSymbol{\delta}{\mathalpha}{egreeki}{"0E}
\DeclareMathSymbol{\epsilon}{\mathalpha}{egreeki}{"0F}
\DeclareMathSymbol{\zeta}{\mathalpha}{egreeki}{"10}
\DeclareMathSymbol{\eta}{\mathalpha}{egreeki}{"11}
\DeclareMathSymbol{\theta}{\mathalpha}{egreeki}{"12}
\DeclareMathSymbol{\iota}{\mathalpha}{egreeki}{"13}
\DeclareMathSymbol{\kappa}{\mathalpha}{egreeki}{"14}
\DeclareMathSymbol{\lambda}{\mathalpha}{egreeki}{"15}
\DeclareMathSymbol{\mu}{\mathalpha}{egreeki}{"16}
\DeclareMathSymbol{\nu}{\mathalpha}{egreeki}{"17}
\DeclareMathSymbol{\xi}{\mathalpha}{egreeki}{"18}
\DeclareMathSymbol{\omicron}{\mathalpha}{egreeki}{"6F}
\DeclareMathSymbol{\pi}{\mathalpha}{egreeki}{"19}
\DeclareMathSymbol{\rho}{\mathalpha}{egreeki}{"1A}
\DeclareMathSymbol{\sigma}{\mathalpha}{egreeki}{"1B}
 \DeclareMathSymbol{\tau}{\mathalpha}{egreeki}{"1C}
\DeclareMathSymbol{\upsilon}{\mathalpha}{egreeki}{"1D}
\DeclareMathSymbol{\phi}{\mathalpha}{egreeki}{"1E}
\DeclareMathSymbol{\chi}{\mathalpha}{egreeki}{"1F}
\DeclareMathSymbol{\psi}{\mathalpha}{egreeki}{"20}
\DeclareMathSymbol{\omega}{\mathalpha}{egreeki}{"21}
\DeclareMathSymbol{\varepsilon}{\mathalpha}{egreeki}{"22}
\DeclareMathSymbol{\vartheta}{\mathalpha}{egreeki}{"23}
\DeclareMathSymbol{\varpi}{\mathalpha}{egreeki}{"24}
\let\varrho\rho 
\let\varsigma\sigma
 \let\varkappa\kappa
\DeclareMathSymbol{\varphi}{\mathalpha}{egreeki}{"27}
\DeclareMathSymbol{\varAlpha}{\mathalpha}{egreeki}{"41}
\DeclareMathSymbol{\varBeta}{\mathalpha}{egreeki}{"42}
\DeclareMathSymbol{\varGamma}{\mathalpha}{egreeki}{"00}
\DeclareMathSymbol{\varDelta}{\mathalpha}{egreeki}{"01}
\DeclareMathSymbol{\varEpsilon}{\mathalpha}{egreeki}{"45}
\DeclareMathSymbol{\varZeta}{\mathalpha}{egreeki}{"5A}
\DeclareMathSymbol{\varEta}{\mathalpha}{egreeki}{"48}
\DeclareMathSymbol{\varTheta}{\mathalpha}{egreeki}{"02}
 \DeclareMathSymbol{\varIota}{\mathalpha}{egreeki}{"49}
\DeclareMathSymbol{\varKappa}{\mathalpha}{egreeki}{"4B}
\DeclareMathSymbol{\varLambda}{\mathalpha}{egreeki}{"03}
\DeclareMathSymbol{\varMu}{\mathalpha}{egreeki}{"4D}
\DeclareMathSymbol{\varNu}{\mathalpha}{egreeki}{"4E}
\DeclareMathSymbol{\varXi}{\mathalpha}{egreeki}{"04}
\DeclareMathSymbol{\varOmicron}{\mathalpha}{egreeki}{"4F}
\DeclareMathSymbol{\varPi}{\mathalpha}{egreeki}{"05}
\DeclareMathSymbol{\varRho}{\mathalpha}{egreeki}{"50}
\DeclareMathSymbol{\varSigma}{\mathalpha}{egreeki}{"06}
\DeclareMathSymbol{\varTau}{\mathalpha}{egreeki}{"54}
\DeclareMathSymbol{\varUpsilon}{\mathalpha}{egreeki}{"07}
\DeclareMathSymbol{\varPhi}{\mathalpha}{egreeki}{"08}
\DeclareMathSymbol{\varChi}{\mathalpha}{egreeki}{"58}
\DeclareMathSymbol{\varPsi}{\mathalpha}{egreeki}{"09}
\DeclareMathSymbol{\varOmega}{\mathalpha}{egreeki}{"0A} 
\DeclareMathSymbol{\Alpha}{\mathalpha}{egreekr}{"41}
\DeclareMathSymbol{\Beta}{\mathalpha}{egreekr}{"42}
\DeclareMathSymbol{\Gamma}{\mathalpha}{egreekr}{"00}
\DeclareMathSymbol{\Delta}{\mathalpha}{egreekr}{"01}
\DeclareMathSymbol{\Epsilon}{\mathalpha}{egreekr}{"45}
\DeclareMathSymbol{\Zeta}{\mathalpha}{egreekr}{"5A}
\DeclareMathSymbol{\Eta}{\mathalpha}{egreekr}{"48}
\DeclareMathSymbol{\Theta}{\mathalpha}{egreekr}{"02}
\DeclareMathSymbol{\Iota}{\mathalpha}{egreekr}{"49}
\DeclareMathSymbol{\Kappa}{\mathalpha}{egreekr}{"4B}
\DeclareMathSymbol{\Lambda}{\mathalpha}{egreekr}{"03}
\DeclareMathSymbol{\Mu}{\mathalpha}{egreekr}{"4D}
\DeclareMathSymbol{\Nu}{\mathalpha}{egreekr}{"4E}
\DeclareMathSymbol{\Xi}{\mathalpha}{egreekr}{"04}
\DeclareMathSymbol{\Omicron}{\mathalpha}{egreekr}{"4F}
\DeclareMathSymbol{\Pi}{\mathalpha}{egreekr}{"05}
\DeclareMathSymbol{\Rho}{\mathalpha}{egreekr}{"50}
\DeclareMathSymbol{\Sigma}{\mathalpha}{egreekr}{"06}
\DeclareMathSymbol{\Tau}{\mathalpha}{egreekr}{"54}
\DeclareMathSymbol{\Upsilon}{\mathalpha}{egreekr}{"07}
\DeclareMathSymbol{\Phi}{\mathalpha}{egreekr}{"08}
\DeclareMathSymbol{\Chi}{\mathalpha}{egreekr}{"58}
\DeclareMathSymbol{\Psi}{\mathalpha}{egreekr}{"09}
\DeclareMathSymbol{\Omega}{\mathalpha}{egreekr}{"0A}
\DeclareMathSymbol{\alphaup}{\mathalpha}{egreekr}{"0B}
\DeclareMathSymbol{\betaup}{\mathalpha}{egreekr}{"0C}
\DeclareMathSymbol{\gammaup}{\mathalpha}{egreekr}{"0D}
 \DeclareMathSymbol{\deltaup}{\mathalpha}{egreekr}{"0E}
\DeclareMathSymbol{\epsilonup}{\mathalpha}{egreekr}{"0F}
\DeclareMathSymbol{\zetaup}{\mathalpha}{egreekr}{"10}
\DeclareMathSymbol{\etaup}{\mathalpha}{egreekr}{"11}
\DeclareMathSymbol{\thetaup}{\mathalpha}{egreekr}{"12}
\DeclareMathSymbol{\iotaup}{\mathalpha}{egreekr}{"13}
\DeclareMathSymbol{\kappaup}{\mathalpha}{egreekr}{"14}
\DeclareMathSymbol{\lambdaup}{\mathalpha}{egreekr}{"15}
\DeclareMathSymbol{\muup}{\mathalpha}{egreekr}{"16}
\DeclareMathSymbol{\nuup}{\mathalpha}{egreekr}{"17}
\DeclareMathSymbol{\xiup}{\mathalpha}{egreekr}{"18}
\DeclareMathSymbol{\omicronup}{\mathalpha}{egreekr}{"6F}
  \DeclareMathSymbol{\piup}{\mathalpha}{egreekr}{"19}
\DeclareMathSymbol{\rhoup}{\mathalpha}{egreekr}{"1A}
\DeclareMathSymbol{\sigmaup}{\mathalpha}{egreekr}{"1B}
\DeclareMathSymbol{\tauup}{\mathalpha}{egreekr}{"1C}
\DeclareMathSymbol{\upsilonup}{\mathalpha}{egreekr}{"1D}
\DeclareMathSymbol{\phiup}{\mathalpha}{egreekr}{"1E}
\DeclareMathSymbol{\chiup}{\mathalpha}{egreekr}{"1F}
\DeclareMathSymbol{\psiup}{\mathalpha}{egreekr}{"20}
\DeclareMathSymbol{\omegaup}{\mathalpha}{egreekr}{"21}
\DeclareMathSymbol{\varepsilonup}{\mathalpha}{egreekr}{"22}
\DeclareMathSymbol{\varthetaup}{\mathalpha}{egreekr}{"23}
\DeclareMathSymbol{\varpiup}{\mathalpha}{egreekr}{"24}

\DeclareMathSymbol{\varphiup}{\mathalpha}{egreekr}{"27}

\renewcommand\sfdefault{uop}
\DeclareMathAlphabet{\mathsf}  {T1}{\sfdefault}{m}{sl}
\SetMathAlphabet{\mathsf}{bold}{T1}{\sfdefault}{b}{sl}
\newcommand*{\mathte}[1]{\textbf{\textit{\textsf{#1}}}}

\usepackage[scaled=0.84]{DejaVuSansMono}

\usepackage{mathdots}

\usepackage[usenames]{xcolor}
\definecolor{mypurpleblue}{RGB}{68,119,170}
\definecolor{myblue}{RGB}{102,204,238}
\definecolor{mygreen}{RGB}{34,136,51}
\definecolor{myyellow}{RGB}{204,187,68}
\definecolor{myred}{RGB}{238,102,119}
\definecolor{myredpurple}{RGB}{170,51,119}
\definecolor{mygrey}{RGB}{187,187,187}
\definecolor{lgrey}{RGB}{221,221,221}
\colorlet{shadecolor}{lgrey}

\usepackage{bm}

\usepackage{microtype}

\usepackage[backend=biber,mcite,
citestyle=authoryear-comp,bibstyle=pglpm-authoryear,autopunct=false,sorting=ny,sortcites=false,natbib=false,maxcitenames=2,maxbibnames=8,minbibnames=8,giveninits=true,uniquename=false,uniquelist=false,maxalphanames=1,block=space,hyperref=true,defernumbers=false,useprefix=true,sortupper=false,language=british,parentracker=false,autocite=footnote]{biblatex}
\DeclareSortingTemplate{ny}{\sort{\field{sortname}\field{author}\field{editor}}\sort{\field{year}}}
\DeclareFieldFormat{postnote}{#1}
\DefineBibliographyExtras{british}{\def\finalandcomma{\addcomma}}
\renewcommand*{\finalnamedelim}{\addspace\amp\space}
\DeclareDelimFormat{multicitedelim}{\addsemicolon\addspace\space}
\DeclareDelimFormat{compcitedelim}{\addsemicolon\addspace\space}
\DeclareDelimFormat{postnotedelim}{\addspace}
\ifarxiv\else\fi
\renewcommand{\bibfont}{\footnotesize}
\defbibheading{bibliography}[\bibname]{\section*{#1}\addcontentsline{toc}{section}{#1}
}
\newcommand*{\citep}{\footcites}

\providecommand{\href}[2]{#2}

\newcommand*{\amp}{\&}

\newcommand*{\subtitleproc}[1]{}

\def\myUrlOrds{\do\0\do\1\do\2\do\3\do\4\do\5\do\6\do\7\do\8\do\9\do\a\do\b\do\c\do\d\do\e\do\f\do\g\do\h\do\i\do\j\do\k\do\l\do\m\do\n\do\o\do\p\do\q\do\r\do\s\do\t\do\u\do\v\do\w\do\x\do\y\do\z\do\A\do\B\do\C\do\D\do\E\do\F\do\G\do\H\do\I\do\J\do\K\do\L\do\M\do\N\do\O\do\P\do\Q\do\R\do\S\do\T\do\U\do\V\do\W\do\X\do\Y\do\Z}%
\makeatletter
\g@addto@macro{\UrlBreaks}{\myUrlOrds}
\makeatother
\newcommand*{\arxiveprint}[1]{%
arXiv \doi{10.48550/arXiv.#1}%
}
\newcommand*{\mparceprint}[1]{%
\href{http://www.ma.utexas.edu/mp_arc-bin/mpa?yn=#1}{mp\_arc:\allowbreak\nolinkurl{#1}}%
}
\newcommand*{\haleprint}[1]{%
\href{https://hal.archives-ouvertes.fr/#1}{\textsc{hal}:\allowbreak\nolinkurl{#1}}%
}
\newcommand*{\philscieprint}[1]{%
\href{http://philsci-archive.pitt.edu/archive/#1}{PhilSci:\allowbreak\nolinkurl{#1}}%
}
\newcommand*{\doi}[1]{%
\href{https://doi.org/#1}{\textsc{doi}:\allowbreak\nolinkurl{#1}}%
}
\newcommand*{\biorxiveprint}[1]{%
bioRxiv \doi{10.1101/#1}%
}
\newcommand*{\osfeprint}[1]{%
Open Science Framework \doi{10.31219/osf.io/#1}%
}
\newcommand*{\osfproj}[1]{%
Open Science Framework \doi{10.17605/osf.io/#1}%
}

\usepackage{graphicx}

\PassOptionsToPackage{hyphens}{url}\usepackage[hypertexnames=false,pdfencoding=unicode,psdextra]{hyperref}

\usepackage[depth=4]{bookmark}
\hypersetup{colorlinks=true,bookmarksnumbered,pdfborder={0 0 0.25},citebordercolor={0.2667 0.4667 0.6667},citecolor=mypurpleblue,linkbordercolor={0.6667 0.2 0.4667},linkcolor=myredpurple,urlbordercolor={0.1333 0.5333 0.2},urlcolor=mygreen,breaklinks=true,pdftitle={\pdftitle},pdfauthor={\pdfauthor}}

\usepackage[british]{datetime2}
\DTMnewdatestyle{mydate}%
{
}
\DTMsetdatestyle{mydate}

\ifafour\setstocksize{297mm}{210mm}
\else\setstocksize{210mm}{5.5in}
\fi
\settrimmedsize{\stockheight}{\stockwidth}{*}
\setlxvchars[\normalfont] 
\setxlvchars[\normalfont]
\ifafour\settypeblocksize{*}{32pc}{1.618} 
\else\settypeblocksize{*}{26pc}{1.618}
\fi
\setulmargins{*}{*}{1}
\setlrmargins{*}{*}{*}
\setheadfoot{\onelineskip}{2.5\onelineskip}
\setheaderspaces{*}{2\onelineskip}{*}
\setmarginnotes{2ex}{10mm}{0pt}
\checkandfixthelayout[nearest]

\newenvironment{acknowledgements}{\section*{Thanks}\addcontentsline{toc}{section}{Thanks}}{\par}
\newenvironment{contributions}{\section*{Author contributions}\addcontentsline{toc}{section}{Author contributions}}{\par}
\makeatletter\renewcommand{\appendix}{\par
  \bigskip{\centering
   \interlinepenalty \@M
   \normalfont
   \printchaptertitle{\sffamily\appendixpagename}\par}
  \setcounter{section}{0}%
  \gdef\@chapapp{\appendixname}%
  \gdef\thesection{\@Alph\c@section}%
  \anappendixtrue}\makeatother
\counterwithout{section}{chapter}
\setsecnumformat{\upshape\csname the#1\endcsname\quad}
\setsecheadstyle{\large\bfseries\sffamily%
\centering}
\setsubsecheadstyle{\bfseries\sffamily%
\raggedright}
\setsubsecindent{0pt}
\setparaheadstyle{\bfseries\sffamily%
\raggedright}
\newcommand{\addchap}[1]{\chapter*[#1]{#1}\addcontentsline{toc}{chapter}{#1}}
\newcommand{\addsec}[1]{\section*{#1}\addcontentsline{toc}{section}{#1}}

\copypagestyle{manaart}{plain}
\makeheadrule{manaart}{\headwidth}{0.5\normalrulethickness}
\makeoddhead{manaart}{%
{\footnotesize
\scshape\headauthor}}{}{{\footnotesize\sffamily%
\headtitle}}
\makeoddfoot{manaart}{}{\thepage}{}

\definecolor{mygray}{gray}{0.333}
\iftypodisclaim%
\ifafour\newcommand\addprintnote{\begin{picture}(0,0)%
\put(245,149){\makebox(0,0){\rotatebox{90}{\tiny\color{mygray}\textsf{This
            document is designed for screen reading and
            two-up printing on A4 or Letter paper}}}}%
\end{picture}}
\else\newcommand\addprintnote{\begin{picture}(0,0)%
\put(176,112){\makebox(0,0){\rotatebox{90}{\tiny\color{mygray}\textsf{This
            document is designed for screen reading and
            two-up printing on A4 or Letter paper}}}}%
\end{picture}}\fi
\makeoddfoot{plain}{}{\makebox[0pt]{\thepage}\addprintnote}{}
\else
\makeoddfoot{plain}{}{\makebox[0pt]{\thepage}}{}
\fi
\makeoddhead{plain}{\scriptsize\reporthead}{}{}

\pretitle{\begin{center}\LARGE\sffamily%
\bfseries}
\posttitle{\bigskip\end{center}}

\makeatletter\newcommand*{\atf}{\includegraphics[totalheight=\heightof{@}]{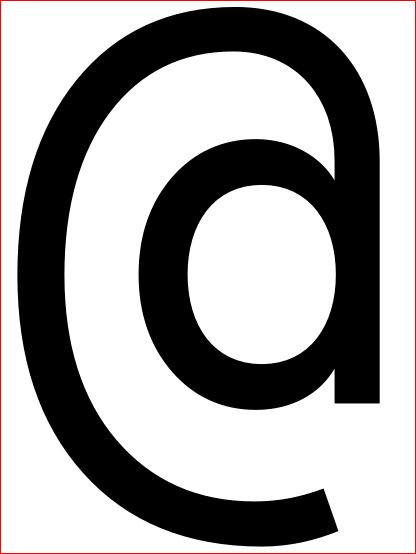}}\makeatother

\providecommand{\epost}[1]{\texttt{\footnotesize\textless#1\textgreater}}
\providecommand{\email}[2]{\href{mailto:#1ZZ@#2 ((remove ZZ))}{#1\protect\atf#2}}

\preauthor{\vspace{-0\baselineskip}\begin{center}
\normalsize\sffamily%
\lineskip  0.5em}
\postauthor{\par\end{center}}
\predate{\DTMsetdatestyle{mydate}\begin{center}\footnotesize}
\postdate{\end{center}\vspace{-\medskipamount}}

\setfloatadjustment{figure}{\footnotesize}
\captiondelim{\quad}
\captionnamefont{\footnotesize\sffamily%
}
\captiontitlefont{\footnotesize}
\midsloppy
\paragraphfootnotes
\footmarkstyle{\textsuperscript{
\scriptsize\bfseries#1}~}

\definecolor{notecolour}{RGB}{68,170,153}

\title{\propertitle}
\author{%
  K. Dyrland \href{https://orcid.org/0000-0002-7674-5733}{\protect\includegraphics[scale=0.16]{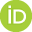}}\\[-\jot]
  {\scriptsize\epost{\email{kjetil.dyrland}{gmail.com}}}%
\\[\jot]%
A. S. Lundervold \href{https://orcid.org/0000-0001-8663-4247}{\protect\includegraphics[scale=0.16]{orcid_32x32.png}}\textsuperscript{\ensuremath{\dagger}} \\[-\jot]
{\scriptsize\epost{\email{alexander.selvikvag.lundervold}{hvl.no}}}%
\\[\jot]%
P.G.L.  Porta Mana  \href{https://orcid.org/0000-0002-6070-0784}{\protect\includegraphics[scale=0.16]{orcid_32x32.png}}\\[-\jot]
{\scriptsize\epost{\email{pgl}{portamana.org}}}%
\\{\tiny(listed alphabetically)}
\\[\jot]
{\footnotesize Dept of Computer science, Electrical Engineering and Mathematical Sciences,\\
  Western Norway University of Applied Sciences, Bergen, Norway
  \\[\jot]
  \textsuperscript{\ensuremath{\dagger}}\amp\ Mohn Medical Imaging and Visualization Centre, Dept of Radiology,\\
  Haukeland University Hospital, Bergen, Norway

}
}

\date{\firstpublished; updated \updated}

\newcommand*{\delt}{\deltaup}
\newcommand*{\I}{\mathrm{i}}
\newcommand*{\e}{\mathrm{e}}
\newcommand*{\di}{\mathrm{d}}
\newcommand*{\defd}{\coloneqq}

\newcommand*{\dotv}{\mathord{\,\cdot\,}}
\renewcommand*{\le}{\leqslant}
\renewcommand*{\ge}{\geqslant}
\DeclarePairedDelimiter\clcl{[}{]}
\DeclarePairedDelimiter\abs{\lvert}{\rvert}
\DeclarePairedDelimiter\set{\{}{\}} 
\newcommand*{\p}{\mathrm{p}}
\renewcommand*{\P}{\mathrm{P}}
\newcommand*{\E}{\mathrm{E}}

\newcommand*{\sect}{\S}
\newcommand*{\chap}{ch.}%
\newcommand*{\chaps}{chs}%
\newcommand*{\eqn}{eq.}%
\newcommand*{\fig}{fig.}%
\newcommand*{\figs}{figs}%

\newcommand*{\eg}{{e.g.}}

\newcommand*{\cf}{{cf.}}
\newcommand*{\etal}{{et al.}}

\newcommand*{\myhat}[1]{{\mkern1.5mu\skew{2}\hat{\mkern-1.5mu#1\mkern-1.5mu}\mkern 1.5mu}}

\renewcommand*{\texteuro}{\$}

\DeclareMathOperator*{\argmax}{arg\,max}
\newcommand*{\ml}{machine-learning}
\newcommand*{\itemyes}{{\fontencoding{U}\fontfamily{pzd}\selectfont\symbol{51}}}
\newcommand*{\itemno}{{\fontencoding{U}\fontfamily{pzd}\selectfont\symbol{55}}}
\newcommand*{\good}[1]{\ensuremath{{\color{mypurpleblue}\bm{#1}}}}
\newcommand*{\bad}[1]{\ensuremath{{\color{myredpurple}#1}}}
\newcommand*{\cx}{X}
\newcommand*{\cy}{Y}
\newcommand*{\eu}{\bar{U}}
\newcommand*{\aveu}{\myhat{\mathte{U}}}
\newcommand*{\uncu}[1]{\mathte{U}^{(#1)}}
\newcommand*{\umatrix}[4]{\begin{bmatrix*}[r]#1&#2\\#3&#4\end{bmatrix*}}
\newcommand*{\sumatrix}[4]{\begin{bsmallmatrix*}[r]#1&#2\\#3&#4\end{bsmallmatrix*}}
\newcounter{dummy}

\begin{document}
\captiondelim{\quad}\captionnamefont{\footnotesize}\captiontitlefont{\footnotesize}
\selectlanguage{british}\frenchspacing
\maketitle

\abstractrunin
\abslabeldelim{}
\renewcommand*{\abstractname}{}
\renewcommand*{\abstracttextfont}{\normalfont\footnotesize}
\setlength{\abstitleskip}{-\absparindent}
\begin{abstract}\labelsep 0pt%
  \noindent 
  How can one meaningfully make a measurement, if the meter does not conform to any standard and its scale expands or shrinks depending on what is measured? In the present work it is argued that current evaluation practices for \ml\ classifiers are affected by this kind of problem, leading to negative consequences when classifiers are put to real use; consequences that could have been avoided. It is proposed that evaluation be grounded on Decision Theory, and the implications of such foundation are explored. The main result is that every evaluation metric must be a linear combination of confusion-matrix elements, with coefficients -- \enquote{utilities} -- that depend on the specific classification problem. For binary classification, the space of such possible metrics is effectively two-dimensional. It is shown that popular metrics such as precision, balanced accuracy, Matthews Correlation Coefficient, Fowlkes-Mallows index, $F_{1}$-measure, and Area Under the Curve are never optimal: they always give rise to an in-principle \emph{avoidable} fraction of incorrect evaluations. This fraction is even larger than would be caused by the use of a decision-theoretic metric with moderately wrong coefficients.


\end{abstract}
\selectlanguage{british}\frenchspacing



\setcounter{section}{-1}

\section{Prologue: a short story}
\label{sec:intro}

The manager of a factory which produces a sort of electronic component wishes to employ a \ml\ classifier to assess the durability of each produced component. The durability determines whether the component will be used in one of two possible kinds of device. The classifier should take some complex features of the component as input, and output one of the two labels \enquote{0} for \enquote{long durability}, or \enquote{1} for \enquote{short durability}, depending on the component type.

Two candidate classifiers, let us call them $\Alpha$ and $\Beta$, are trained on available training data. When employed on a separate evaluation set, they yield the following confusion matrices, written in the format
\begin{equation*}
  \rotatebox[origin=c]{90}{
    \clap{\textit{\parbox{5em}{\centering\scriptsize classifier\\output\\$1\quad 0$}}
    }}\ 
    \overbracket[0pt]{
      \umatrix{
    \text{\footnotesize True 0} }{ \text{\footnotesize False 0}}{
    \text{\footnotesize False 1} }{ \text{\footnotesize True 1}}
      }^{
      \clap{\textit{\parbox{6em}{\centering\scriptsize true class\\$0\hspace{3em}1$}}
    }}
\end{equation*}
and normalized over the total number of evaluation data:
\begin{align}
  \label{eq:CM_A}
\text{classifier $\Alpha$:}\quad  \umatrix{
    0.27 }{ 0.15 }{ 0.23 }{ 0.35
  }
\ ,
  \\
  \label{eq:CM_B}
\text{classifier $\Beta$:}\quad  \umatrix{
    0.43 }{ 0.18 }{ 0.07 }{ 0.32
  }
\ .
\end{align}
These matrices show that the factory produces, on average, 50\% short- and 50\% long-durability components.

The confusion matrices above lead to the following values of common evaluation metrics\autocites[Balanced accuracy:][]{brodersenetal2010}[$F_{1}$ measure:][]{vanrijsbergen1974}[Matthews correlation coefficient:][]{matthews1975}[Fowlkes-Mallows index:][]{fowlkesetal1983} for the two classifiers. Class~$0$ is \enquote{positive}, $1$ \enquote{negative}. \textbf{\color{mypurpleblue}Blue bold} indicates the classifier favoured by the metric, {\color{myredpurple}red} the disfavoured:
\begin{table}[!h]\centering\footnotesize
  \caption{}\label{tab:example_metrics}
  \begin{tabular}{lcc}
    Metric & classifier $\Alpha$ & classifier $\Beta$\\
    \hline
    Accuracy (also balanced accuracy) & \bad{0.62} & \good{0.75} \\
    Precision & \bad{0.64} & \good{0.70} \\
    $F_{1}$ measure & \bad{0.59} & \good{0.77} \\
    Matthews Correlation Coefficient & \bad{0.24} & \good{0.51} \\
    Fowlkes-Mallows index & \bad{0.59} & \good{0.78} \\
    True-positive rate (recall) & \bad{0.54} & \good{0.86} \\
    True-negative rate (specificity) & \good{0.70} & \bad{0.64}
  \end{tabular}
\end{table}\FloatBlock
The majority of these metrics favour classifier $\Beta$, some of them by quite a wide relative difference. Only the true-negative rate favours classifier $\Alpha$, but only by a relative difference of 9\%. 

The developers of the classifiers therefore recommend the employment of classifier $\Beta$.

The factory manager does not fully trust these metrics, asking, \enquote*{how do I know they are appropriate?}. The developers assure that these metrics are widely used. The manager (of engineering background) comments, \enquote*{I don't remember `widely used' being a criterion of scientific correctness -- not after Galileo at least}, and decides to employ both classifiers for a trial period, to see which factually leads to the best revenue. The two classifiers are integrated into two separate but otherwise identical parallel production lines.

During the trial period, the classifiers perform according to the classification statistics of the confusion matrices~\eqref{eq:CM_A} and \eqref{eq:CM_B} above. At the end of this period the factory manager finds that the average net gains per assessed component yielded by the two classifiers are\footnote{\enquote{\texteuro} represents a generic currency or value unit; this is why it is not written in front of the gains.}
\begin{equation}
  \label{eq:final_gains}
  \begin{tabular}{cc}
 classifier $\Alpha$ & classifier $\Beta$\\
    \hline
    \good{3.5}\,\text{\texteuro} & \bad{-3.5}\,\text{\texteuro}
 \end{tabular}
\end{equation}
That is, classifier $\Beta$ actually led to a \emph{loss} of revenue. The manager therefore decides to employ classifier $\Alpha$, commenting with a smug smile that it is always unwise to trust the recommendations of developers, unacquainted with the nitty-gritty reality of a business.

The average gains above are easy to calculate from some additional information. The final net gains caused by the correct or incorrect classification of one electronic component are as follows:
\begin{equation}
  \label{eq:utility_example}
  \rotatebox[origin=c]{90}{
    \clap{\textit{\parbox{5em}{\centering\scriptsize classifier\\output\\$1\quad 0$}}
    }}\ 
    \overbracket[0pt]{
      \umatrix{
        15\,\text{\texteuro} }{ -335\,\text{\texteuro} }{
        -35\,\text{\texteuro} }{ 165\,\text{\texteuro}}
      }^{
      \clap{\textit{\parbox{6em}{\centering\scriptsize true class\\$0\hspace{3em}1$}}
    }}
\end{equation}
The reason behind these values is that short-durability components (class~1) provide more power and are used in high-end, costly devices; but they cause extreme damage and consequent repair costs and refunds if used in devices that require long-durability components (class~0). Long-durability components provide less power and are used in low-end, cheaper devices; they cause some damage if used in devices that require short-durability components, but with lower consequent costs.

Taking the sum of the products of the gains above by the respective percentages of occurrence -- that is, the elements of the confusion matrix -- yields the final average gain. The final average gain returned by the use of classifier~$\Alpha$, for example, is
\begin{equation*}
  15\,\text{\texteuro} \times 0.27 
  -335\,\text{\texteuro} \times 0.15 
  -35\,\text{\texteuro} \times 0.23 
  + 165\,\text{\texteuro} \times 0.35 =
  3.5 \,\text{\texteuro} \ .
\end{equation*}
In the present case, the confusion matrices~\eqref{eq:CM_A} and \eqref{eq:CM_B} lead to the amounts \eqref{eq:final_gains} found by the manager.




\section{Issues in the evaluation of classifiers}
\label{sec:issues}

The story above illustrates several well-known issues of currently popular evaluation procedures for \ml\ classifiers:
\begin{enumerate}
\item We are swept by an avalanche of possible evaluation metrics. Often it is not clear which is the most compelling. In the story above, for example, one could argue that the true-negative rate was the appropriate metric, in view of the great difference in gains between correct and wrong classification for class~1, compared with that for class~0.
But at which point does this qualitative reasoning fail? Imagine that the net gains had been as follows instead:
\begin{equation}
  \label{eq:utility_example_2}
  \rotatebox[origin=c]{90}{
    \clap{\textit{\parbox{5em}{\centering\scriptsize classifier\\output\\$1\hspace{1.5em}0$}}
    }}\ 
  \overbracket[0pt]{
    \umatrix{
        45\,\text{\texteuro} }{ -335\,\text{\texteuro} }{
        -65\,\text{\texteuro} }{ 165\,\text{\texteuro}
      }}^{
      \clap{\textit{\parbox{6em}{\centering\scriptsize true class\\$0\hspace{4em}1$}}
    }} \ .
\end{equation}
Also in this case one could argue that there is a greater economic difference between correct and wrong classification for class~1 than for class~0. The true-negative rate should, therefore, again be the appropriate metric. Yet a simple calculation analogous to the one of \sect~\ref{sec:intro} shows that classifier~$\Beta$ actually leads to the best average revenue: $7.3\,\text{\texteuro/component}$, vs $4.7\,\text{\texteuro/component}$ for classifier $\Alpha$. Hence the true-negative rate is \emph{not} the appropriate metric in this case: our qualitative reasoning failed us.

\item A classifier favoured by the majority of available metrics can still turn out \emph{not} to be the best one in practice.

\item\label{item:ad_hoc} Most popular metrics are introduced by intuitive reasoning, ad hoc mathematical operations, special assumptions (such as Gaussianity\autocites[e.g.][\sect~31 p.~183 for the Matthews correlation coefficient]{fisher1925_r1963} or other statistical assumptions), and an analysis of special cases only. Unfortunately this kind of derivations does not guarantee generalization to all cases, nor that the proposed metric is uniquely determined by the chosen assumptions, nor that it satisfies more general consistency requirements.

  By contrast, consider the kind of derivation that starts from specific qualitative requirements and mathematically proves the \emph{uniqueness} of a particular formula satisfying them. Examples are the derivation of the Shannon entropy  as the \emph{unique} metric universally satisfying a set of basic requirements for the amount of information \autocites{shannon1948}[\sect~3.2]{woodward1953_r1964}[also][]{goodetal1968}. Or  the derivation  of the probability calculus as the \emph{unique} set of rules satisfying general rational requirements for inductive reasoning, learning, and prediction\footnote{\cites{cox1946,fine1973,halpern1999b,snow1998,snow2001}[\chaps~1--2]{jaynes1994_r2003}; see also \cites{selfetal1987,cheeseman1988}[\chap~12]{russelletal1995_r2022}.}.
Or the derivation of decision theory as the unique framework guaranteeing a rational and optimal decision under uncertainty \autocites[\sect~15.2]{russelletal1995_r2022}[\chaps~2--3]{vonneumannetal1944_r1955}.
  
\item\label{item:hope_medical} Let us assume that some of the popular metrics identify the best algorithm \enquote{in the majority of cases} -- although it is difficult to statistically define such a majority, and no real surveys have ever been conducted to back up this assumption. Yet, do we expect the end-user to simply \emph{hope} not to belong to the unlucky minority? Is such uncertainty inevitable?

  We cannot have a cavalier attitude towards this problem: life and death can depend on it in some \ml\ applications \autocites[cf.][]{howard1980}. Imagine a story analogous to the factory one, but in a medical setting instead. The classifiers should distinguish between two tumour types, requiring two different types of medical intervention. The confusion matrices are the same~\eqref{eq:CM_A} and \eqref{eq:CM_B}. Correct and incorrect classification lead to the following expected remaining life lengths for patients in a specific age range: \autocites[\cf\ the discussion in][\sect~11.2.9]{soxetal1988_r2013}
\begin{equation}
  \label{eq:utility_example_medicine}
  \rotatebox[origin=c]{90}{
    \clap{\textit{\parbox{5em}{\centering\scriptsize classifier\\output\\$1\hspace{1.5em}0$}}
    }}\ 
    \overbracket[0pt]{
      \umatrix{
        350\,\text{months} }{ 0\,\text{months}  }{
        300\,\text{months} }{ 500\,\text{months}
      }}^{
      \clap{\textit{\parbox{8em}{\scriptsize\centering true class\\$0\hspace{7em}1$}}
    }} \ .
\end{equation}
These values might arise in several scenarios. For example, tumours of class~0 and~1 may require very different kinds of treatment. If a class~0 tumour is misdiagnosed and not properly treated, it leads to immediate death (0\,months); if correctly diagnosed, its treatment is usually successful, leading to high life expectancy (500\,months). Class~0 tumours can be treated, but they lead to a shorter life expectancy (350\,months). If they are misdiagnosed as class~1, however, the damage caused by class~1 treatment shortens this life expectancy even further (300\,months).

This matrix above is numerically equivalent to~\eqref{eq:utility_example} up to a common additive constant of $335$, so the final net gains are also shifted by this amount. It is easy to see that the metrics are exactly as in Table~\ref{tab:example_metrics}, the majority favouring classifier~$\Beta$. And yet the use of classifier~$\Alpha$ leads to a more than six-month longer expected remaining life than classifier~$\Beta$.

\item Often it is not possible to temporarily deploy all candidate classifiers, as our fictitious manager did, in order to observe which factually leads to the best results. Or it may even be unethical: consider a situation like the medical one above, where a classifier may lead to a larger number of immediate deaths than another.

\item Finally, all issues listed above are not caused by class imbalance (the occurrence of one class with a higher frequency than another). In our story, for example, the two classes were perfectly balanced. Class imbalance can make all these issues worse \autocites{jenietal2013,zhu2020}.

\end{enumerate}

\bigskip

But our story also points to a possible solution for all these issues. The \enquote{metric} that ultimately proved to be relevant to the manager was the average net monetary gain obtained by using a candidate classifier. In the medical variation discussed in issue~\ref{item:hope_medical} above, it was the average life expectancy. In either case, such metric could have been easily calculated beforehand, upon gathering information about the average gains and losses of correct and incorrect classification, collected in the matrix~\eqref{eq:utility_example} or~\eqref{eq:utility_example_medicine}, and combining these with statistics collected in the confusion matrix associated with the classifier. Denoting the former kind of matrix by $(U_{ij})$ and the confusion matrix by $(C_{ij})$, where $i$ indexes the classifier outputs (rows) and $j$ the true classes (columns), such a metric would have the formula
\begin{equation}
  \label{eq:expected_utility}
  \sum_{i,j} U_{ij}\ C_{ij} 
\end{equation}
the sum extending to all matrix elements.

\medskip

In the present work, we argue that formula~\eqref{eq:expected_utility} is indeed the only acceptable metric for evaluating and comparing the performance of two or more classifiers, each with its own confusion matrix $(C_{ij})$ collected on relevant test data. The coefficients $U_{ij}$, called \emph{utilities}, are problem-dependent. This formula is the \emph{utility yield} of a classifier having confusion matrix $(C_{ij})$.

Our argument is based on \emph{Decision Theory}, an overview of which is given in \sect~\ref{sec:decision_theory}.

The utility yield~\eqref{eq:expected_utility} is a linear combination of the confusion-matrix elements, with coefficients independent of the elements themselves. In \sect~\ref{sec:evaluation_metrics} we explore some properties of this formula and of the space of such metrics for binary classification. We also show that some common metrics such as precision, $F_{1}$-measure, Matthews correlation coefficient, balanced accuracy, and Fowlkes-Mallows index \emph{cannot} be written as a linear combination of this kind (or a one-one function thereof). This impossibility has two consequences. First, it means that these metrics are likely affected by some kind of cognitive bias. Second, there exists \emph{no} classification problem for which these metrics can correctly rank the performance of all pairs of classifiers. Using any one of these metrics leaves open the possibility that the evaluation is incorrect \emph{a priori}. In \sect~\ref{sec:auc} we show that this is also true for the Area Under the Curve of the Receiver Operating Characteristic, and we offer some additional remarks about it from the standpoint of decision theory.

On the other hand, metrics such as accuracy, true-positive rate, true-negative rate can be written in the form~\eqref{eq:expected_utility}. Consequently, each one has a set of classification problems in which it correctly ranks the performance of \emph{all} pairs of classifiers.

What happens if we are uncertain about the utilities appropriate to a classification problem? And what happens if the utilities are incorrectly assessed? We show in \sect~\ref{sec:unknown_wrong_utilities} that uncertainty about utilities still leads to a metric of the form~\eqref{eq:expected_utility}. We also show that an evaluation using incorrect utilities, even with relative errors as large as 20\% of the maximal utility, still leads to a higher amount of correctly ranked classifiers than the use of any of the popular metrics mentioned above.

We summarize and discuss our results in the final \sect~\ref{sec:discussion}.

\section{Brief overview of decision theory}
\label{sec:decision_theory}

\subsection{References}
\label{sec:dt_refs}

Here we give a brief overview of decision theory. We only focus on the notions relevant to the problem of evaluating classifiers, and simply state the rules of the theory. These rules are quite intuitive, but it must be remarked that they are constructed in order to be logically and mathematically self-consistent: see the following references. For a presentation of decision theory from the point of view of artificial intelligence and machine learning, see \cite[\chap~15]{russelletal1995_r2022}. Simple introductions are given by \cites{north1968,raiffa1968_r1970,lindley1971_r1988}[\chap~8]{tribus1969}{jeffrey1965}; and a discussion of its foundations and history by \cite{steeleetal2015_r2020}. For more thorough expositions see \cite{raiffaetal1961_r2000,fentonetal2019,berger1980_r1985,savage1954_r1972}; and \cite{soxetal1988_r2013,huninketal2001_r2014} for a medical perspective. See also Ramsey's \cite*{ramsey1926} insightful and charming pioneering discussion.

\subsection{Decisions and classes}
\label{sec:dt_dec_classes}

Decision theory makes a distinction between
\begin{itemize}
\item the possible situations we are uncertain about: in our case, the possible classes;
\item the possible decisions we can make.
\end{itemize}
This distinction is important because it prevents the appearance of various cognitive biases \autocites{kahnemanetal1982_r2008,gilovichetal2002_r2009,kahneman2011} in evaluating the probabilities and frequencies of the possible situations on the one hand, and the values of our decisions on the other. Examples are the scarcity bias\autocites{camereretal1989,kimetal1999,mittoneetal2009} \enquote*{this class is rare, \emph{therefore} its correct classification must lead to high gains}, and plain wishful thinking: \enquote*{this event leads to high gains, \emph{therefore} it is more probable}.

Often even the number of classes and the number of decisions differ. But in using \ml\ classifiers, one typically considers situations where the set of available decisions and the set of possible classes have some kind of natural correspondence and equal cardinality. In a \enquote{cat vs dog} image classification, for example, the classes are \enquote{cat} and \enquote{dog}, and the decisions could be \enquote{put into folder Cats} vs \enquote{put into folder Dogs}. In a medical application the classes could be \enquote{ill} and \enquote{healthy} and the decisions \enquote{treat} vs \enquote{dismiss}. As already mentioned,  most of our discussions and examples focus for simplicity on binary classification.

\subsection{Utilities and maximization of expected utility}
\label{sec:dt_utilities}

To each decision we associate several \emph{utilities}, depending on which of the possible classes is actually true. A utility may, for instance, equal a gain or loss in money, energy, number of customers, life expectancy, or quality of life, measured in appropriate units; or it may equal a  combination of such quantities.

These utilities are collected into a \emph{utility matrix} $(U_{ij})$, like the ones shown in formulae~\eqref{eq:utility_example}, \eqref{eq:utility_example_2}, \eqref{eq:utility_example_medicine}.
The component $U_{ij}$ is the utility of the decision corresponding to class $i$ if class $j$ is true, or simply the utility of class $i$ conditional on class $j$.

In an individual classification instance, if we know which class is true, then the optimal decision is the one having maximal utility among those conditional on the true class. If, on the other hand, we are uncertain about which class is true, with probability $p_{j}$ for class $j$ such that $\sum_{j}p_{j}=1$, then decision theory states that the optimal decision is the one having maximal \emph{expected} utility $\eu_{i}$, defined as the expected value of the utility of decision $i$ with respect to the probabilities of the various classes:
\begin{equation}
  \label{eq:exp_utility}
  \eu_{i} \defd \sum_{j} U_{ij}\ p_{j} \ .
\end{equation}
In formulae, this principle of \emph{maximization of expected utility}  is
\begin{equation}
  \label{eq:max_expe_utility}
  \text{choose class}\quad
  i^{*} = \argmax_{i}\set{\eu_{i}} \equiv \argmax_{i}\set[\bigg]{\sum_{j} U_{ij}\ p_{j}} \ .
\end{equation}


A very important result in decision theory is that basic requirements of rational decision-making imply that there \emph{must} be a set of utilities underlying the decisions of a rational agent, and the decisions must obey the principle of maximization of expected utility\autocites[\sect~15.2]{russelletal1995_r2022}[\chaps~2--3]{vonneumannetal1944_r1955}.

How are utilities determined? They are obviously problem-specific and cannot be given by the theory (which would otherwise be a model rather than a theory). Utilities can be obvious in decision problems involving gains or losses of measurable quantities such as money or energy (the utility of money is usually not equal to the amount of money, the relationship between the two being somewhat logarithmic \autocites[\eg][pp.~203--204]{north1968}[\chap~4]{raiffa1968_r1970}). In medical problems they can correspond to life expectancy and quality of life; see for example \cite[esp. \chap~8 and \sect~11.2.9]{soxetal1988_r2013} and \cite[esp. \chap~4]{huninketal2001_r2014} on how such health factors are transformed into utilities.

The final utility of a single classification instance may depend, in some cases, on a sequence of further uncertain events and further decisions. In the story of \sect~\ref{sec:intro}, for instance, the misclassification of a short-durability component as a long-durability one leads the final device to break only in a high fraction of cases, and in such cases the end customer requires a refund in a high fraction of subcases; the refunded amount may even depend on further circumstances. The negative utility $U_{01} = -335\,\text{\texteuro}$ in table~\eqref{eq:utility_example} comes from a statistical average of the losses in all these possible end results. This is the topic of so-called decision networks or influence diagrams \autocites[Besides the general references already given:][\sect~15.5]{russelletal1995_r2022}{howardetal1984b_r2005}[for a step-by-step tutorial:][]{raiffa1968_r1970}. The decision-theory subfield of \emph{utility theory} gives rules that guarantee the mutual consistency of a set of utilities in single decisions or decision networks. For simple introductions to utility theory see \cite[\sect~15.2]{russelletal1995_r2022}, \cite[pp.~201--205]{north1968}, and the references given at the beginning of the present section.

In the present work, we do not worry about such rules in order not to complicate the discussion: they should be approximately satisfied if the utilities of a problem have been carefully assessed.

\section{Evaluation of classifiers from a decision-theoretic perspective}
\label{sec:evaluation_metrics}

\subsection{Admissible evaluation metrics for classification problems}
\label{sec:admissible_metrics}

Maximization of expected utility is the ground rule for rational decision making \autocites[We discuss and use it in our companion work][]{dyrlandetal2022b}. In the present work we focus on the stage where a large number of classifications have already been made by a classifier on a test dataset with $N$ data. Denote by $F_{ij}$ the number of instances in which the classifier chose class $i$ and the true class was $j$. Then $(F_{ij})$ is the confusion matrix of the classifier on this particular test set. For all instances in which the classifier chose class $i$ and the true class was $j$, a utility $U_{ij}$ is eventually gained. The total utility yielded by the classifier on the test set is therefore $\sum_{ij} U_{ij}\ F_{ij}$. Dividing by $N$ we obtain the average utility per datum, which we call the \emph{utility yield}; it can be written as
\begin{empheq}[box=\widefbox]{equation}
  \label{eq:final_utility}
  \sum_{ij} U_{ij}\ C_{ij}
\end{empheq}
where $C_{ij} \defd F_{ij}/N$ is the relative frequency of choice $i$ and true class $j$, and $(C_{ij})$ is the normalized confusion matrix.

\emph{The utility yield, formula~\eqref{eq:final_utility}, is therefore the natural metric to evaluate and compare the performance of classifiers on a test set for a classification problem characterized by the utility matrix $(U_{ij})$}.

Note how the utilities $U_{ij}$ do depend on the frequencies $F_{ij}$ or $C_{ij}$. If they did, it would mean that we had waited until \emph{all} classification instances had been made in order to assess the value of each \emph{single} instance. In virtually all classification problem we can think of, this would be a source of evaluation bias, such as the scarcity bias mentioned in \sect~\ref{sec:dt_dec_classes}. It would, moreover, be an impossible procedure in contexts where the consequence of a single classification is manifest before the next classification is made.

If we modify the elements of a utility matrix by a common additive constant or by a common positive multiplicative constant,
\begin{equation}
  \label{eq:modify_UM}
  U_{ij} \mapsto a\ U_{ij} + b  \qquad a > 0 \ ,
\end{equation}
then the final utilities yielded by a classifier with a particular confusion matrix are modified by the same constants. The ranking of any set of classifiers will therefore be the same. After all, an additive constant or a positive factor represent only changes in the zero or the measurement unit of our utility scale \autocites[cf.][\sect~15.2.2]{russelletal1995_r2022}. Such changes should not affect a decision problem. Indeed, the fact that they do not is another example of the logical consistency of decision theory.

\subsection{Space of utility matrices for binary classification}
\label{sec:dt_space_util}

Let us consider a problem of binary classification. It is characterized by a matrix of $2 \times 2$ utilities. We suppose that they are not all equal;  the choice of class would be immaterial otherwise, and the classification problem trivial. We can use the freedom of choosing a zero and measurement unit to bring the utility matrix to a standard form. Let us choose them such that the maximum utility is $1$ and the minimum utility is $0$ (note that this value may still correspond to an actual monetary loss, for example). That is, we are effecting the transformation
\begin{equation}
  \label{eq:normalize_utilities}
  U_{ij} \mapsto \frac{U_{ij} - \min(U_{ij})}{\max(U_{ij}) - \min(U_{ij})} \ .
\end{equation}
With this convention, it is clear that we only have two degrees of freedom in choosing the utility matrix of a binary-classification problem. As a consequence, \emph{the space (more precisely: manifold) of possible evaluation metrics for binary classifications is two-dimensional}. In order to evaluate candidate classifiers for a binary-classification problem, we must choose a point from this space.

\begin{figure}[t]
  \centering
  \includegraphics[width=0.65\linewidth]{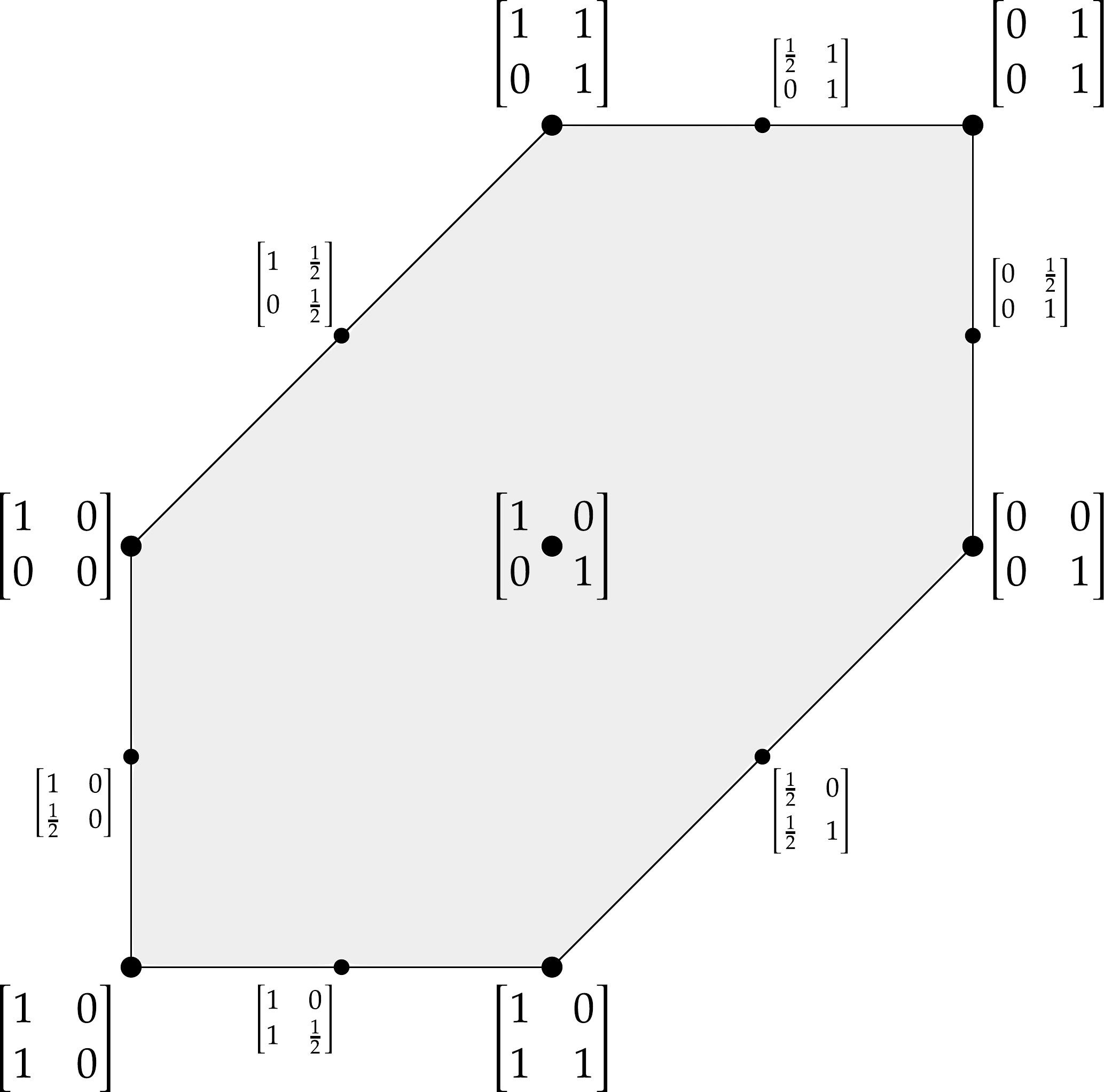}\\
  \caption{Space of utility matrices for binary classification.}
  \label{fig:space_UM}
\end{figure}
We can represent this space as in \fig~\ref{fig:space_UM}. The centre is the utility matrix with equal maximum utilities for correct classification and equal minimum utilities for incorrect classification; we shall see later that it corresponds to the use of accuracy as the evaluation metric. Moving to the left from the centre, the utility for correct classification of class~1 decreases with respect to class~0; vice versa moving to the right. Moving upwards from the centre, the utility for misclassification of class~1 increases; moving downwards, the utility for misclassification of class~0 increases. We have excluded utility matrices in which misclassification has a higher utility than correct classification (although they may occur in some situations); they would appear in the missing upper-left and lower-right corners. Fixing $(x,y)$ axes through the centre of the set, a utility matrix has coordinates
\begin{equation}
  \label{eq:coords_um}
  \begin{bmatrix}
    1 - x\ \delt(x > 0) & y\ \delt(y>0) \\
    -y\ \delt(y<0) & 1 + x\ \delt(x < 0)
  \end{bmatrix} \ ,
  \qquad \abs{x},\abs{y} \le 1 \ .
\end{equation}

Note that this representation is \emph{not} meant to reflect any convex or metric properties, however. No metric or distance is defined in the space of utility matrices. Convex combination is defined if we drop the normalization~\eqref{eq:normalize_utilities} 
but it is not correctly reflected in the representation of \fig~\ref{fig:space_UM}.

\subsection{Relationship with common metrics}
\label{sec:common_metrics}

In \sect~\ref{sec:admissible_metrics} we found that the most general evaluation metric according to decision theory must be a linear combination of the confusion-matrix elements. The coefficients of this linear combination do not depend on the confusion-matrix elements themselves; such a dependence usually reflects some sort of cognitive bias. Which common popular metrics adhere to this mathematical form? We want to answer this question in the binary-classification case while giving as much allowance as possible in the typical context in which popular metrics are used.

Consider the case in which we are comparing several classifiers \emph{on the same test set}. The number of data $N$ and the relative frequencies $f_{0}, f_{1}$ with which the two classes \enquote{$0$}, \enquote{$1$} occur in the test set are fixed and constant for all classifiers under evaluation.

A classifier yields a normalized confusion matrix $(C_{ij})$ which we write in the format
\begin{equation*}
  \rotatebox[origin=c]{90}{
    \clap{\textit{\parbox{5em}{\centering\scriptsize classifier\\output\\$1\hspace{1.5em}0$}}
    }}\ 
    \overbracket[0pt]{
      \umatrix{ C_{00} }{ C_{01} }{ C_{10} }{ C_{11}
      }}^{
      \clap{\textit{\parbox{6em}{\centering\scriptsize true class\\$0\hspace{3em}1$}}
    }} \ .
\end{equation*}

Owing to the constraints $C_{00} + C_{10} \equiv f_{0}$ and $C_{01} + C_{11} \equiv f_{1}$ we can always make two elements of the confusion matrix appear or disappear from any formula, replacing them with expressions involving the remaining two elements and the class frequencies. To avoid ambiguities in interpreting the functional form of mathematical formulae, let us agree to always express them in terms of $C_{00}$ and $C_{11}$ only, making the replacements $C_{10} = f_{0} - C_{00}$, $C_{01} = f_{1} - C_{11}$ wherever necessary.

Recall that, given a utility matrix, we can always modify its elements by a common positive multiplicative constant $a$ and by a common additive constant $b$, \eqn~\eqref{eq:modify_UM}, because such a modification corresponds to a change of unit and zero of the utility scale. With such a modification the evaluation metric~\eqref{eq:final_utility} takes the equivalent form
\begin{equation}
  \label{eq:final_utility_modified}
 a\ \sum_{ij} \ U_{ij}\ C_{ij} + b
\end{equation}
because $\sum_{ij} C_{ij} \equiv 1$. Writing the sum explicitly and rewriting the elements $C_{10}, C_{01}$ in terms of $C_{00}, C_{11}$ as discussed above, this formula becomes 
\begin{equation}
  \label{eq:final_utility_binary}
    a\ (U_{00} - U_{10})\ C_{00} \ + \ 
    a\ (U_{11} - U_{01})\ C_{11} \ + \ 
    a\ f_{0}\ U_{10} + a\ f_{1}\ U_{01} +  b \ .
\end{equation}

Since in the present context $N, f_{0}, f_{1}$ are constants, we are free to construct the arbitrary constants $a > 0$ and $b$ from them in any way we please:
\begin{equation}
  \label{eq:constants_functions}
  a = a(N, f_{0}, f_{1}) > 0\ , \qquad
  b = b(N, f_{0}, f_{1}) \ .
\end{equation}
We can also use this freedom to include the term $a\ f_{0}\ U_{10} + a\ f_{1}\ U_{01}$ into $b$ in the formula above. We conclude that \emph{an evaluation metric for binary classification complies with decision theory if and only if it can be written in the general form}
\begin{equation}
  \label{eq:general_valuation_metric}
  a(N, f_{0}, f_{1})\ \cx\ C_{00} +
  a(N, f_{0}, f_{1})\ \cy\  C_{11} +
  b(N, f_{0}, f_{1})
\end{equation}
\emph{where $\cx,\cy$ are real constants that do not depend on $C_{00}, C_{11}, N, f_{0}, f_{1}$; and $a(\dotv)>0$, $b(\dotv)$ are arbitrary functions of $N, f_{0}, f_{1}$ only.}

A monotonic function (such as an exponential) of the expression above is also admissible if we only require a comparison score to rank several classifiers from best to worst.

Let us examine some common evaluation metrics for binary classification from this point of view. We write their formulae in terms of $C_{00}, C_{11}$.

The following metrics are particular instances of formula~\eqref{eq:general_valuation_metric}:
\begin{itemize}
\item[\itemyes] \emph{Accuracy:} $C_{00}+C_{11}$. We have $a=1$, $\cx=\cy=1$, $b=0$. Indeed it corresponds to the utility yield based on the identity utility matrix $(U_{ij}) = \sumatrix{ 1}{0}{0}{1 }$ (or equivalently a utility matrix that assigns the same utility to the correct classification of any class, and the same, lower utility to the misclassification of any class).
  
\item[\itemyes] \emph{True-positive rate (recall):} $C_{00}/f_{0}$. Here $a=1/f_{0}$, $\cx=1$, $\cy=0$, $b=0$. It corresponds to using the utility matrix $\sumatrix{ 1}{0}{0}{0 }$. 

\item[\itemyes] \emph{True-negative rate (specificity):} $C_{11}/f_{1}$. Here $a=1/f_{1}$, $\cx=0$, $\cy=1$, $b=0$. It corresponds to using the utility matrix $\sumatrix{ 0}{0}{0}{1 }$. 
\end{itemize}

The following metrics instead \emph{cannot} be written in the form~\eqref{eq:general_valuation_metric}, nor as monotonic functions of that form:
\begin{itemize}
\item[\itemno] \emph{Precision:} $C_{00}/(C_{00}-C_{11}+f_{1})$. Non-linear in $C_{00}, C_{11}$.

\item[\itemno] \emph{$F_{1}$-measure:} $2 C_{00}/(C_{00} - C_{11} + 1)$. Non-linear in $C_{00}, C_{11}$. The same is true for the more general $F_{\beta}$-measures.

\item[\itemno] \emph{Matthews correlation coefficient:} $\frac{
    f_{1}\,C_{00} + f_{0}\,C_{11}}{\sqrt{
      f_{0}\ f_{1}\ (f_{1} + C_{00} - C_{11})\ (f_{0} + C_{11} - C_{00})
    }}$. Non-linear in $C_{00}, C_{11}$.

\item[\itemno] \emph{Fowlkes-Mallows index:} $C_{00}/\sqrt{
      f_{0}\ (f_{1} + C_{00} - C_{11})}$. Non-linear in $C_{00}, C_{11}$.

  \item[\itemno] \emph{Balanced accuracy:} $C_{00}/(2 f_{0}) + C_{11}/(2 f_{1})$. Despite being linear in $C_{00}, C_{11}$ and an average of two metrics (true-positive and true-negative rate) that are instances of formula~\eqref{eq:general_valuation_metric}, it is not an instance of that formula, because the two averaged metrics involve different $a(\dotv)$ functions.
\end{itemize}

We see that many popular evaluation metrics do not comply with the principles of decision theory. Any such metric suffers from two problems.

First, as discussed in \sect~\ref{sec:decision_theory}, the metric involves an interdependence of utilities and classification frequencies, which typically implies some form of cognitive bias\autocites[discuss such biases regarding the $F_{1}$-measure]{handetal2018}.

Second, the ranking of confusion matrices yielded by the metric does not fully agree with that yielded by any utility matrix -- a full agreement would otherwise imply that the metric could be written in the form~\eqref{eq:general_valuation_metric}. Some confusion matrices must therefore be incorrectly ranked. Since any rational classification problem is characterized by some underlying utility matrix, this means that the non-compliant metric will always lead to some wrong evaluations. By contrast, compliant metrics such as the accuracy give completely correct rankings for all pairs of confusion matrices in specific sets of classification problems.

\begin{figure}[p]
  \centering
\includegraphics[width=0.95\linewidth]{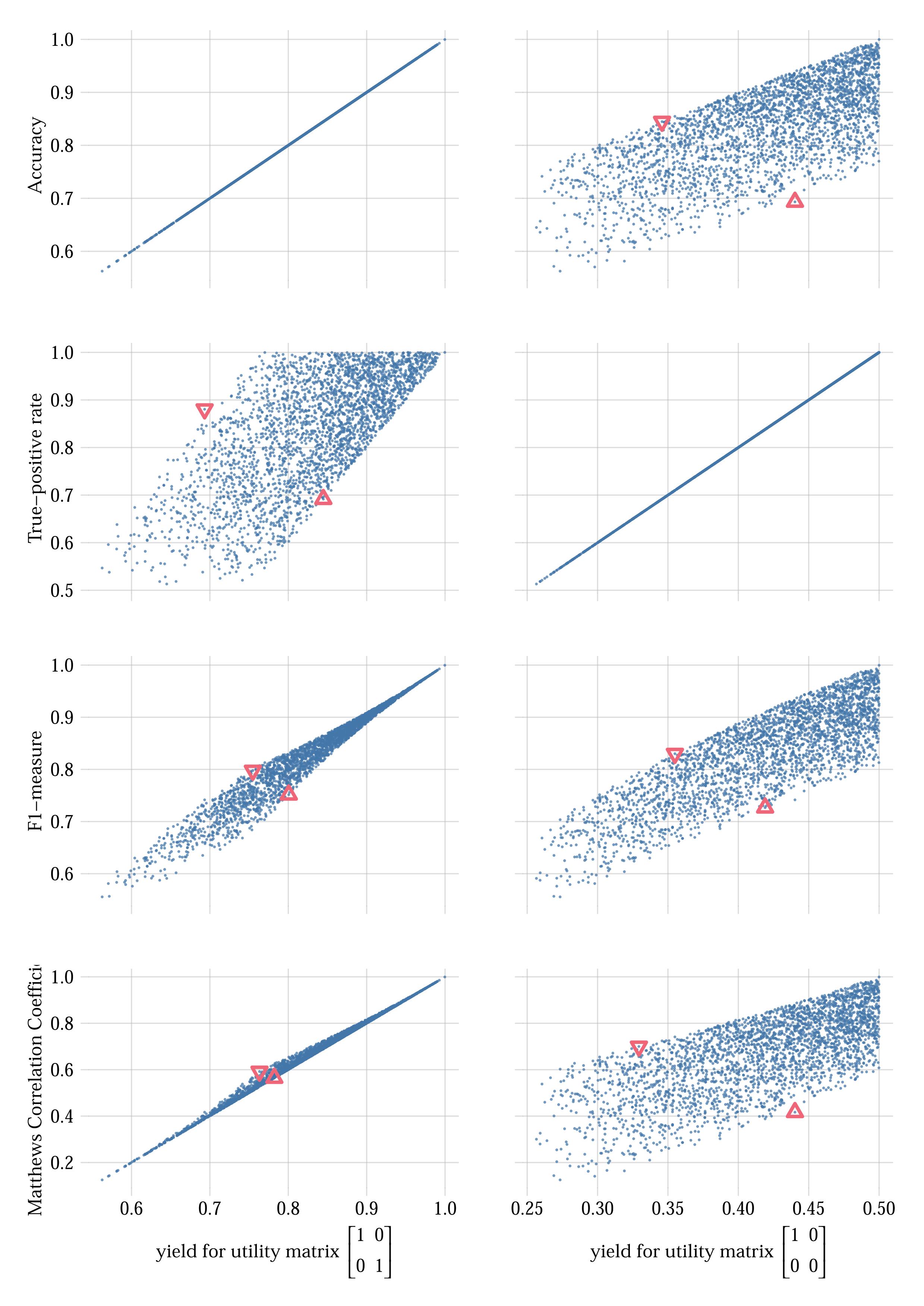}\\
\caption{Relationship between various evaluation metrics and actual utility yields for two different binary-classification problems with underlying utility matrices $\sumatrix{ 1}{0}{0}{1 }$ (left column) and $\sumatrix{ 1}{0}{0}{0 }$ (right column). All confusion matrices ({\color{mypurpleblue}blue dots}) are obtained from a dataset with 50\%/50\% class balance.
  Pairs of {\color{myred}red triangles} in a plot show two confusion matrices that are wrongly ranked by the metric (y-axis) with respect to the actual utility yield (x-axis). Clearly, there can even be three or more confusion matrices ranked in completely reverse order by the metric.
%
  The accuracy gives correct evaluations for the classification problem on the left column; and the true-positive rate, for the one on the right. }
  \label{fig:metrics_vs_utility}
\end{figure}

\begin{figure}[p]
  \centering
\includegraphics[width=0.95\linewidth]{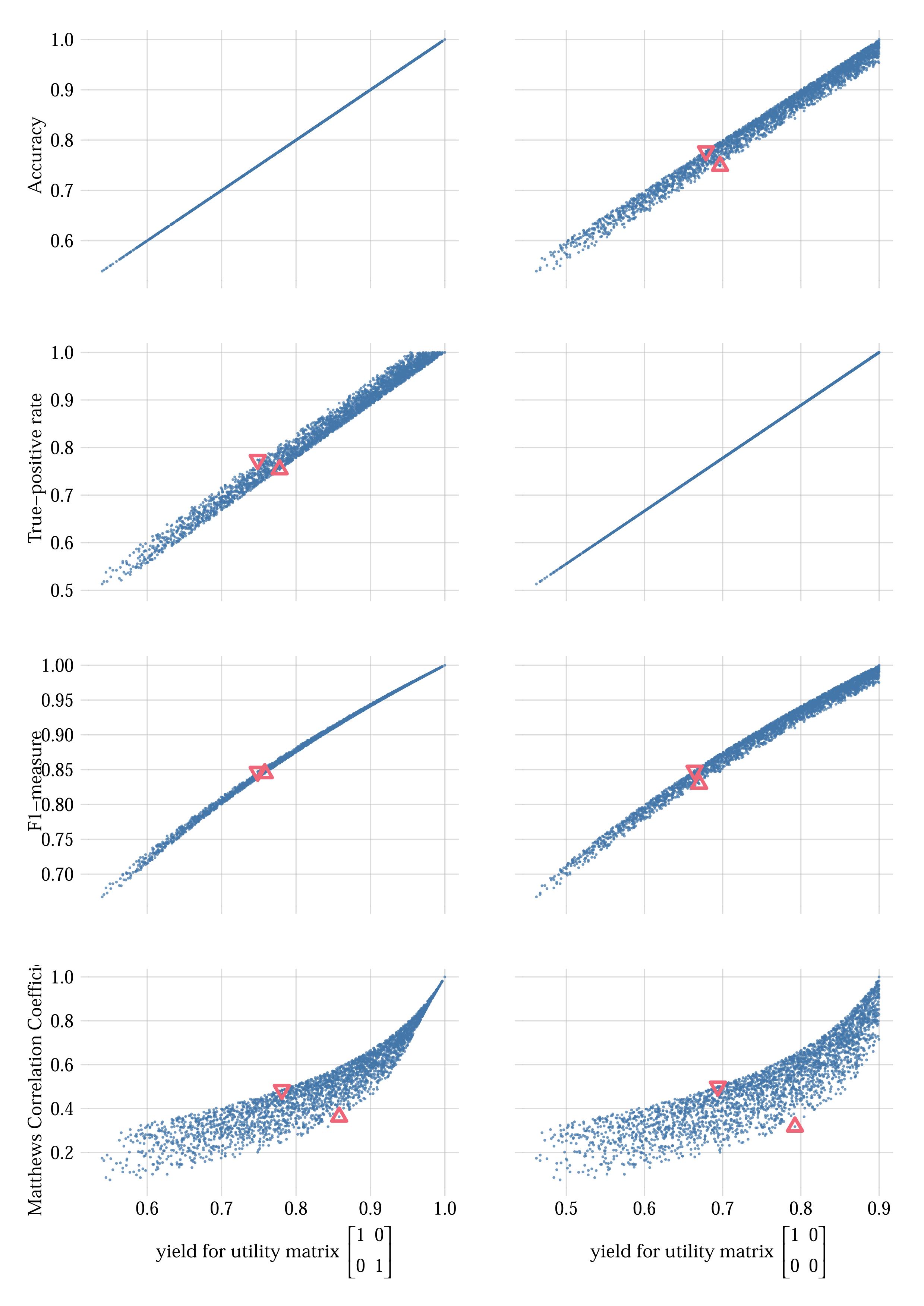}\\
\caption{As for \fig~\ref{fig:metrics_vs_utility} but for confusion matrices obtained from an imbalanced dataset with 90\% occurrence of class~0 (\enquote{positive}) and 10\% of class~1 (\enquote{negative}).}
  \label{fig:metrics_vs_utility2}
\end{figure}

The second phenomenon is illustrated in the plots of \figs~\ref{fig:metrics_vs_utility}--\ref{fig:metrics_vs_utility2}.
Each blue dot in a plot represents a hypothetical confusion matrix obtained from a test dataset in a binary classification. The dot's coordinates are the utility yield of that confusion matrix according to a particular utility matrix underlying the classification problem, and the score of the confusion matrix according to another metric. The underlying utility matrix is $\sumatrix{ 1}{0}{0}{1 }$ for all plots in the left column, and $\sumatrix{ 1}{0}{0}{0 }$ for all plots in the right column. The other metrics considered, one for each row of plots, are accuracy, true-positive rate (recall, class~0 being \enquote{positive}), $F_{1}$-measure, Matthews correlation coefficient.

The confusion matrices are selected by first fixing a proportion of classes in the dataset, which is 50\%/50\% (balanced dataset) for all plots in \fig~\ref{fig:metrics_vs_utility} and 90\%/10\% (imbalanced dataset) for all plots in \fig~\ref{fig:metrics_vs_utility2}. Then a %
true-positive rate and a true-negative rate are independently selected %
\begin{minipage}[t]{0.66\linewidth}
from the range $\clcl{1/2,1}$, with a probability linearly increasing in the rate (median of 0.85, lower and upper quartiles at 0.75 and 0.93; see side plot). These confusion matrices therefore represent the classification statistics produced 
  by classifiers that tend to have high true-negative and true-positive rates -- as is clear from the fact that the points tend to accumulate on the upper-right corners of the plots.
\end{minipage}%
\hfill\begin{minipage}[t]{0.32\linewidth}\vspace{0pt}
\centering\includegraphics[width=\linewidth]{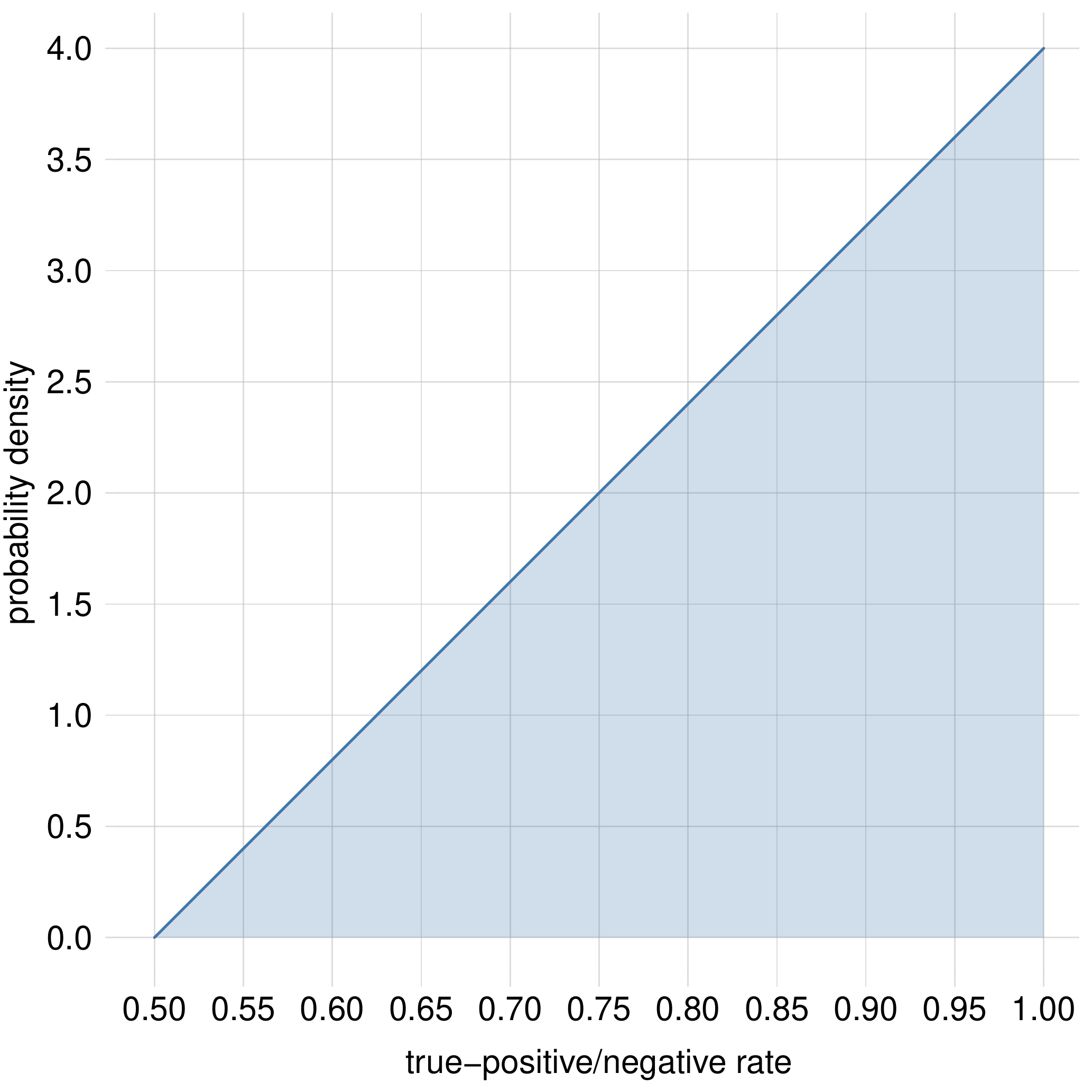}
\refstepcounter{dummy}
\label{fig:linear_distr}  
\end{minipage}

We see that the accuracy (first-row plots) always gives correct relative evaluations of all confusion matrices when the underlying utility matrix is equivalent to $\sumatrix{ 1}{0}{0}{1 }$ (left column): the y-coordinate is a monotonically increasing function -- in fact a linear function -- of the x-coordinate. Accuracy is indeed the utility yield corresponding to the identity utility matrix. The true-positive rate (second-row plots) always gives correct relative evaluations (provided the test set is the same) when the underlying utility matrix is equivalent to $\sumatrix{ 1}{0}{0}{0 }$ (right column).

On the other hand, if any of these two metrics is used for a problem having a different underlying utility matrix, then there is no deterministic relationship between the metric's score and the actual utility yield. In this case it is always possible to find two or more confusion matrices for which the metric gives completely reversed evaluations with respect to the actual utility yield. In other words, the confusion matrix -- and associated algorithm -- which is worst according to the true utility, is ranked best by the metric; and vice versa. Pairs of red triangular shapes in a plot are examples of such confusion matrices wrongly ranked by the y-axis metric.

Metrics such as accuracy and true-positive rate, complying with formula~\eqref{eq:general_valuation_metric}, thus require us to rely on evaluation \emph{luck} only when they are used in the wrong classification problem.

The plots for the $F_{1}$-measure (third-row plots) and Matthews correlation coefficient (fourth-row plots) show that these two metrics do not have any functional relationship with the actual utility yield. It is again always possible to find two or more confusion matrices for which either metric gives completely reversed evaluations with respect to the actual utility yield. But for these two metrics, unlike accuracy and true-positive rate, cases of incorrect evaluation will \emph{always} occur in \emph{every} classification problem.

Metrics such as  $F_{1}$-measure and Matthews correlation coefficient, not complying with formula~\eqref{eq:general_valuation_metric}, thus \emph{always require us to rely on luck in our evaluations}. There are no classification problems for which these metrics lead to always correct evaluations.

\medskip

A metric non-compliant with decision theory can lead to a large number of correct results for some classification problems and test sets. The bottom-left plot of \fig~\ref{fig:metrics_vs_utility}, for instance, shows that the Matthews correlation coefficient is almost a monotonically increasing deterministic function of the utility yield when the underlying utility matrix is the identity and the dataset is balanced (but it is not when the underlying utility matrix is $\sumatrix{ 1}{0}{0}{0 }$ or the dataset is imbalanced; see corresponding plots). Such an occasional partial agreement is useless, however. Knowledge of the utility matrix is a prerequisite for relying on such partial agreement-- but given such knowledge we can directly use the actual utility yield instead, which has an exact agreement and is easier to compute.

\section{Unknown or incorrect utilities}
\label{sec:unknown_wrong_utilities}

So far, we have argued that the natural evaluation metric for a classifier is the utility yield of its confusion matrix, according to the utilities underlying the classification problem of interest. We have also argued that many popular metrics, those not complying with formula~\eqref{eq:general_valuation_metric}, must always a priori lead to instances of incorrect evaluation. Our arguments are based on the principles of decision theory.

Several interrelated questions spring from our arguments, though:
\begin{itemize}
\item What to do when we are uncertain about the utilities underlying a classification problem?
\item What happens if the utilities we use are actually wrong, that is, not the true ones underlying the problem?
\item How often do uncompliant metrics such as $F_{1}$-measure or Matthews correlation coefficient lead to incorrect results, on average?
\end{itemize}
In fact, if a small error in the assessment of the utilities led to a large number of wrong evaluations, while non-compliant metrics led to a small number of wrong evaluations on average, then all the rigorousness of decision-theoretic metrics would be useless in practice, and  non-compliant metrics would be best for real applications.

This is not the case, however. We now discuss how to deal with uncertainty about the utilities and present an important result: Using wrong utilities, even with relative errors almost as large as 20\% of the maximum utility, still leads to fewer incorrect relative evaluations on average than using many currently popular metrics.

\subsection{Unknown utilities; average performance on several classification problems}
\label{sec:unknown_utilities}

Dealing with unknown utilities is straightforward. Suppose we are uncertain whether the utility matrix appropriate to a classification problem is $\uncu{1} \equiv \bigl(U^{(1)}_{ij}\bigr)$, or $\uncu{2}$, or $\uncu{3}$, and so on, where the number of alternatives can even be infinite or continuous. Each alternative $\uncu{a}$ has a probability $q_a$, or probability density $q(a)\ \di a$ in the continuous case. Then \emph{for this classification problem, we should use the expected utility matrix
  \begin{equation}
    \label{eq:expe_UM}
    \aveu \defd q_{1}\ \uncu{1} + q_{2}\ \uncu{2} + q_{3}\ \uncu{3} + \dotsb
  \end{equation}
  or $\aveu \defd \int q(a)\, \uncu{a}\, \di a$ in the continuous case}.

We only give a sketch of the proof of this intuitive result  \autocites[see \eg][esp. \chap~3]{raiffa1968_r1970}. If we are uncertain about the utility matrix, then we have a double decision problem: choosing the optimal utility and choosing the optimal class. If the true utility matrix is, for instance, $\uncu{2} \equiv \bigl(U^{(2)}_{ij}\bigr)$, and the true class is class~$0$, then choosing class~$1$ would yield a utility $U^{(2)}_{10}$, choosing class~$0$ would yield a utility $U^{(2)}_{00}$, and so on. Our double decision problem is thus characterized by a rectangular utility matrix that is the row-concatenation of the utility matrices $\uncu{a}$. We make the realistic judgement that the probabilities $q_{a}$ of the utility matrices and the probabilities $p_{j}$ of the classes are independent, so that $q_{a}\cdot p_{j}$ is the probability that the true utility matrix is $\uncu{a}$ and the true class is $j$. The principle of maximum expected utility, \sect~\ref{sec:dt_utilities} \eqn~\eqref{eq:max_expe_utility}, then leads to the maximization of the expected utilities
\begin{equation}
  \label{eq:average_expe_utility}
  \eu_{i} \defd \sum_{j,a} U^{(a)}_{ij}\ q_{a}\cdot p_{j}
  \equiv \sum_{j} \biggl[\underbrace{\sum_{a} q_{a}\ U^{(a)}_{ij}}_{\aveu} \biggr]\ p_{j}
\end{equation}
in which the expected utility matrix~\eqref{eq:expe_UM} appears as an \enquote{effective} utility matrix to be used for the class-decision problem alone.

If our uncertainty is symmetric with respect to the utilities conditional on the different classes -- for instance, our uncertainty about the utilities conditional on class~$0$ is the same as on class~$1$ -- then the expected utility matrix is equivalent to the identity matrix. The utility yield is in this case equal to the accuracy. The accuracy is therefore the natural evaluation metric to use if we are in a complete state of uncertainty regarding the underlying utilities. This fact is indeed reflected in some results discussed in \sect~\ref{sec:wrong_utility_assess}.

For binary classification the set of possible utility matrices can be represented as in \fig~\ref{fig:space_UM}, as discussed in \sect~\ref{sec:dt_space_util}. Our uncertainty about the true underlying utility matrix corresponds to a discrete or continuous distribution of probability over this set. Note, however, that the expected utility matrix~\eqref{eq:expe_UM} does \emph{not} correspond to the mass-centre of the distribution, because of the peculiar coordinate system used in that figure. The actual mass-centre is obtained by representing the set of utility matrices as a two-dimensional surface (a tetrahedron) in three-dimensional space. For brevity we do not discuss this representation in the present work.

\medskip

The procedure of averaging utilities, formula~\eqref{eq:expe_UM}, also applies if we want to evaluate how a classifier performs on average on several classification problems, which differ in their utility matrices. Again, what we need to use is the average of their utility matrices.

\subsection{Consequences of wrong utility assessments and comparison with common metrics}
\label{sec:wrong_utility_assess}

It may happen that our assessment of the utility matrix of a classification problem is incorrect, especially if it has been made on semi-quantitative grounds owing to a lack of information. Then our comparative evaluations of classifiers may also end up being incorrect. What is the probability of an incorrect comparative evaluation, on average, in such cases? and how does it depend on the amount of error in the utilities? Is it higher than the probability of incorrect evaluation by other metrics?

\begin{figure}[t]
  \centering
  \hspace{\stretch{1}}
  \parbox[t]{0.45\linewidth}{%
    \centering\includegraphics[width=\linewidth]{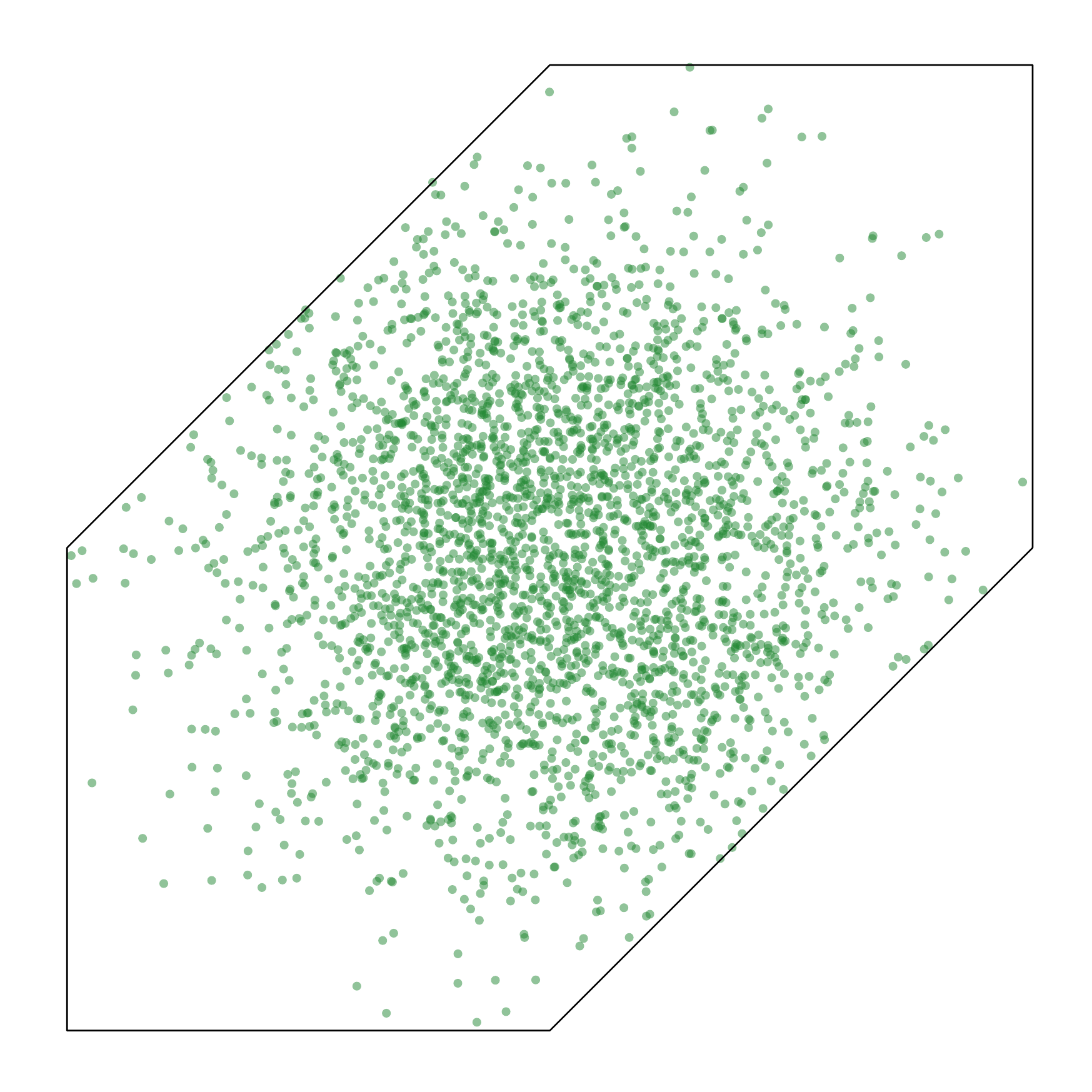}\\
    \caption{Truncated gaussian distribution in the space of utility matrices of \fig~\ref{fig:space_UM}, described in item~\ref{item:distr_um}.}
    \label{fig:gauss_distr_um}}
  \hspace{\stretch{1}}
  \parbox[t]{0.45\linewidth}{%
    \centering\includegraphics[width=\linewidth]{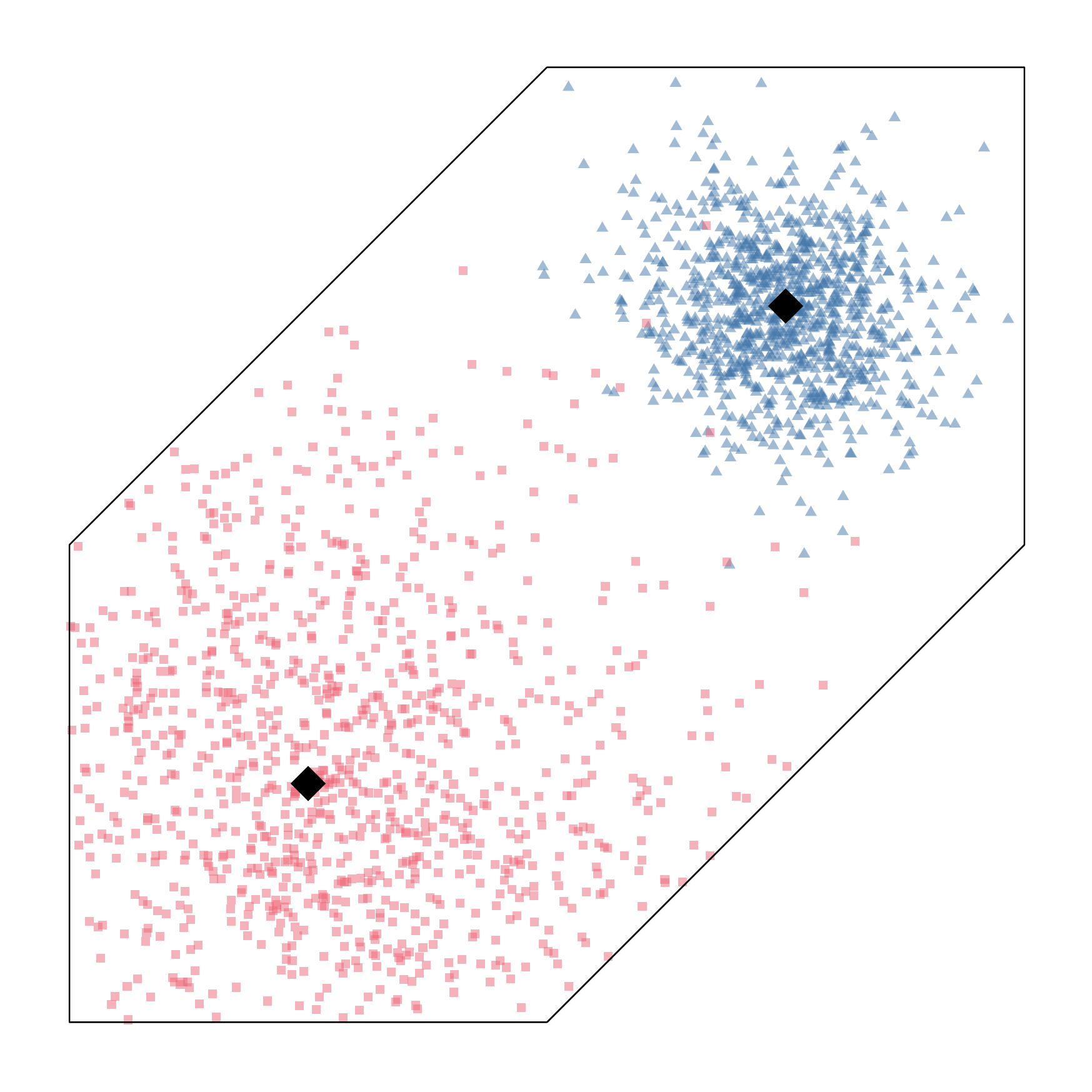}
    \caption{Extents of errors having standard deviations $0.1$ ({\color{mypurpleblue}blue triangles}) and $0.2$ ({\color{myred}red squares}), around the utility matrices $\sumatrix{ 0.5}{0.5}{0}{1 }$ and $\sumatrix{ 1}{0}{0.5}{0.5 }$ (black diamonds).}
    \label{fig:error_distr_um}}
  \hspace{\stretch{1}}
\end{figure}

A precise answer to these questions is extremely difficult if not impossible because to define \enquote{on average} we would need to conduct a survey of classification problems of any kind, collecting statistics about their underlying utility matrices, about the confusion matrices of candidate classification algorithms for their solution, and about the errors committed in assessing utilities. We try to give a cursory answer to the questions above for the binary-classification case, based on the following assumptions and judgements:
\begin{enumerate}[label=(\roman*)]
\item\label{item:distr_um} Two possible distributions of true utility matrices on the set of \fig~\ref{fig:space_UM} (in that coordinate system):   \begin{enumerate*}[label=\arabic*.]
\item a uniform distribution; \item a bivariate (truncated) gaussian distribution centred on the identity matrix $\sumatrix{ 1}{0}{0}{1 }$ and with standard deviation $1/3$ in the $x$ and $y$ coordinates of \eqn~\eqref{eq:coords_um}, illustrated in \fig~\ref{fig:gauss_distr_um}.
\end{enumerate*}

\item\label{item:distr_cm} A distribution of confusion matrices for which the fraction of one class is uniformly distributed in $\clcl{0,1}$, and the true-positive and true-negative rates are independently distributed in $\clcl{0.5, 1}$ with linearly increasing probabilities (median of 0.85, lower and upper quartiles at 0.75 and 0.93; see side plot on p.~\pageref{fig:linear_distr}). This means that we consider problems with highly imbalanced data to be as common as problems with balanced data (a realistic assumption, according to our experience), and candidate classifiers to be generally good.
  
\item\label{item:distr_error} A truncated gaussian distribution of error around each true utility-matrix element, centred on the true utility value. We consider standard deviations ranging from $0$ to $0.3$. The gaussian must be truncated because each true utility has a value between $0$ and $1$, and because we require the utilities of correct classifications to be larger than those of incorrect ones. Figure~\ref{fig:error_distr_um} illustrates the extent of such an error in the space of utility matrices, for standard deviations equal to $0.1$ (blue triangles) and $0.2$ (red squares).
\end{enumerate}

Under these assumptions, we calculate how often a pair of classifiers, having two confusion matrices with the same class proportions, is evaluated in reverse order, with respect to their true utility yield, when an incorrect utility matrix or another metric is used for the evaluation. This calculation is an integration problem that we solve by Monte Carlo sampling. The procedure is intuitive:
\begin{enumerate}[label=\arabic*.,ref=\arabic*]
\item\label{item:draw_um} Select a \enquote{true} utility matrix according to the distribution~\ref{item:distr_um}.
\item\label{item:draw_wrongum} Select errors around the elements of the true utility matrix, according to the distribution~\ref{item:distr_error}, and add them to it.
\item\label{item:draw_cm} Select a class proportion and then two confusion matrices having that class proportion (the class proportion must be the same since the matrices are obtained from the same data), according to the distributions~\ref{item:distr_cm}.
\item\label{item:diff_true} Calculate the signed difference between the true utility yield of the second confusion matrix and that of the first confusion matrix, using the true utility from step~\ref{item:draw_um}. If this difference is positive, then the second confusion matrix has higher utility than the first; if negative, then the first confusion matrix has higher utility than the second.
\item\label{item:diff_other}
    \begin{enumerate}[label=\alph*.]
\item Consider several metrics (precision, Matthews correlation coefficient, and so on). For each, calculate the signed difference between the score it gives to the second confusion matrix, and the score it gives to the first.
\item Consider the erroneous utility matrix from step~\ref{item:draw_wrongum}. Calculate the signed difference between the utility yield of the second confusion matrix and that of the first confusion matrix, using this erroneous utility matrix.
\end{enumerate}
In either case, a positive difference means that the second confusion matrix is ranked \enquote{best} and the second \enquote{worst}, and vice versa for a negative difference.
\item Now go through the signed differences obtained in step~\ref{item:diff_other}, and compare them, in turn, with the signed difference obtained in step~\ref{item:diff_true}. If the difference from step~\ref{item:diff_other} has opposite sign to that of step~\ref{item:diff_true}, then the two confusion matrices are oppositely and incorrectly ranked by the corresponding metric or by the erroneous utility matrix.
\end{enumerate}

\begin{figure}[p]
  \centering
    \includegraphics[width=\linewidth]{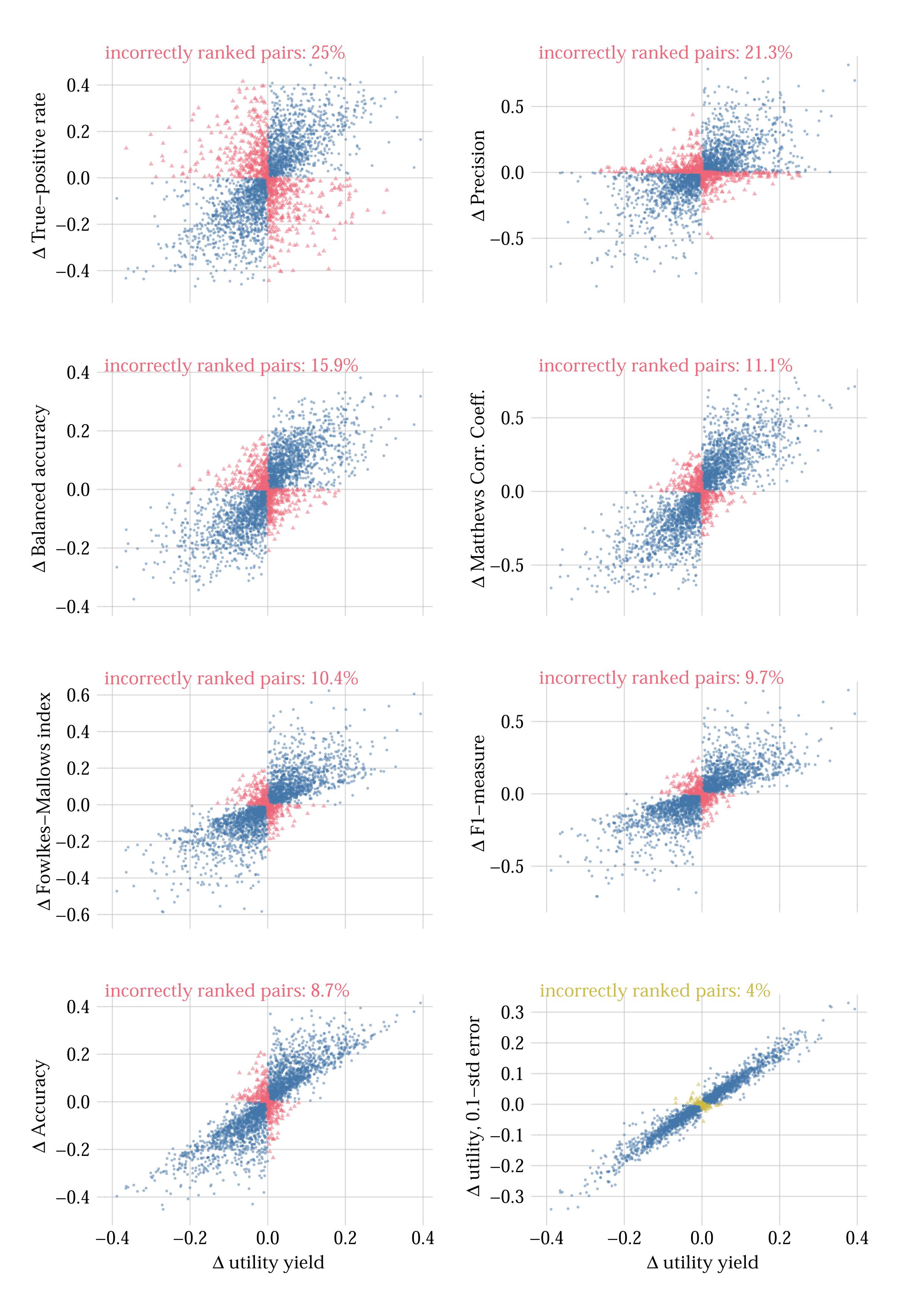}\\
  \caption{Relationship between difference in utility yields according to a \enquote{true} utility matrix, and difference in scores according to other metrics including an incorrectly assessed utility matrix (error with $0.1$ standard deviation). Points landing in the II or IV quadrants represent pairs of confusion matrices that were wrongly compared.}
  \label{fig:wrongly_ranked_pairs}
\end{figure}

The results of this sampling procedure for the case of uniform distribution of true utility matrices, several metrics, and utilities affected by errors with $0.1$ standard deviation, are shown in \fig~\ref{fig:wrongly_ranked_pairs}.
\begin{figure}[p]
  \centering
    \includegraphics[width=0.99\linewidth]{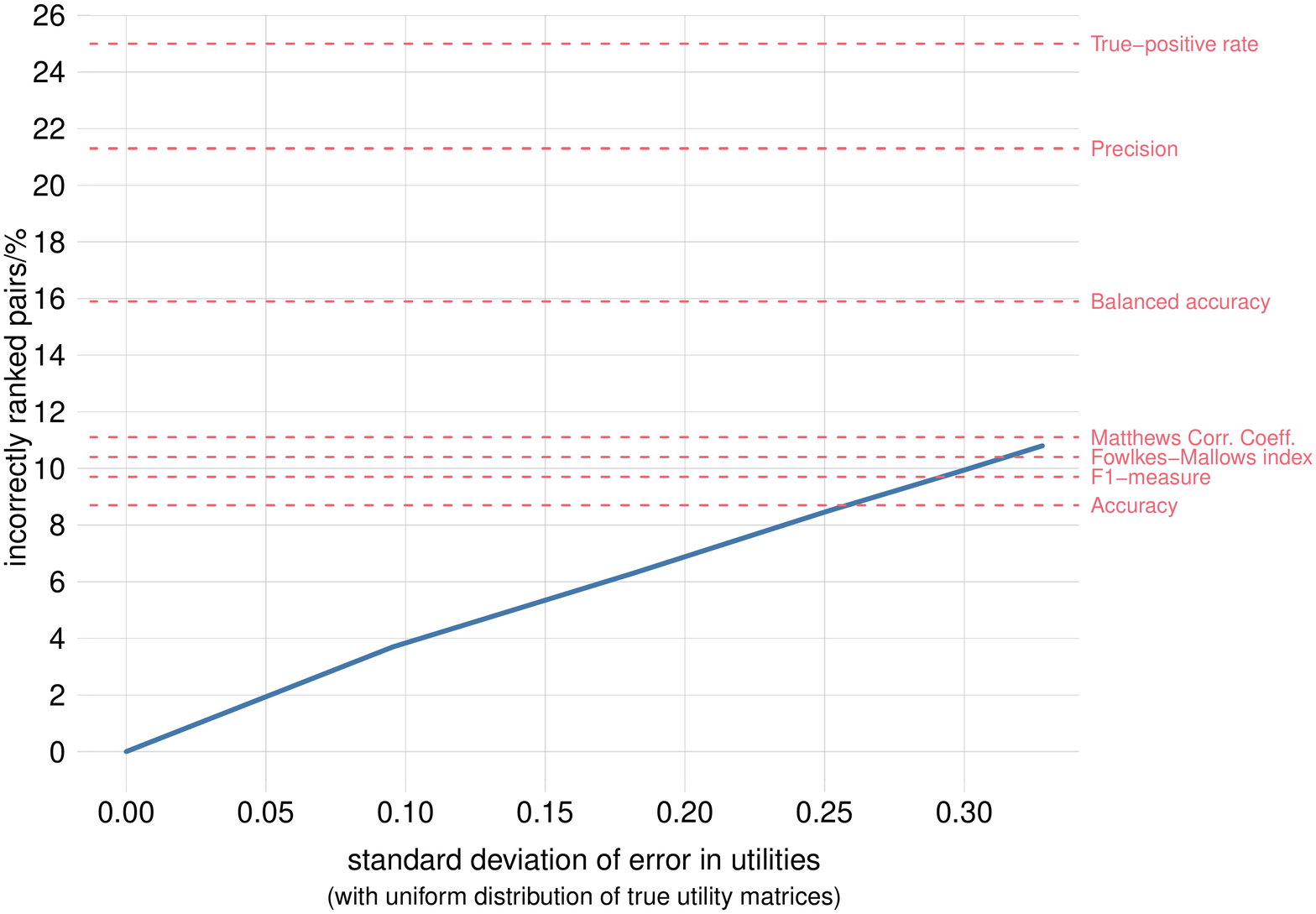}\\\hfill
    \includegraphics[width=0.99\linewidth]{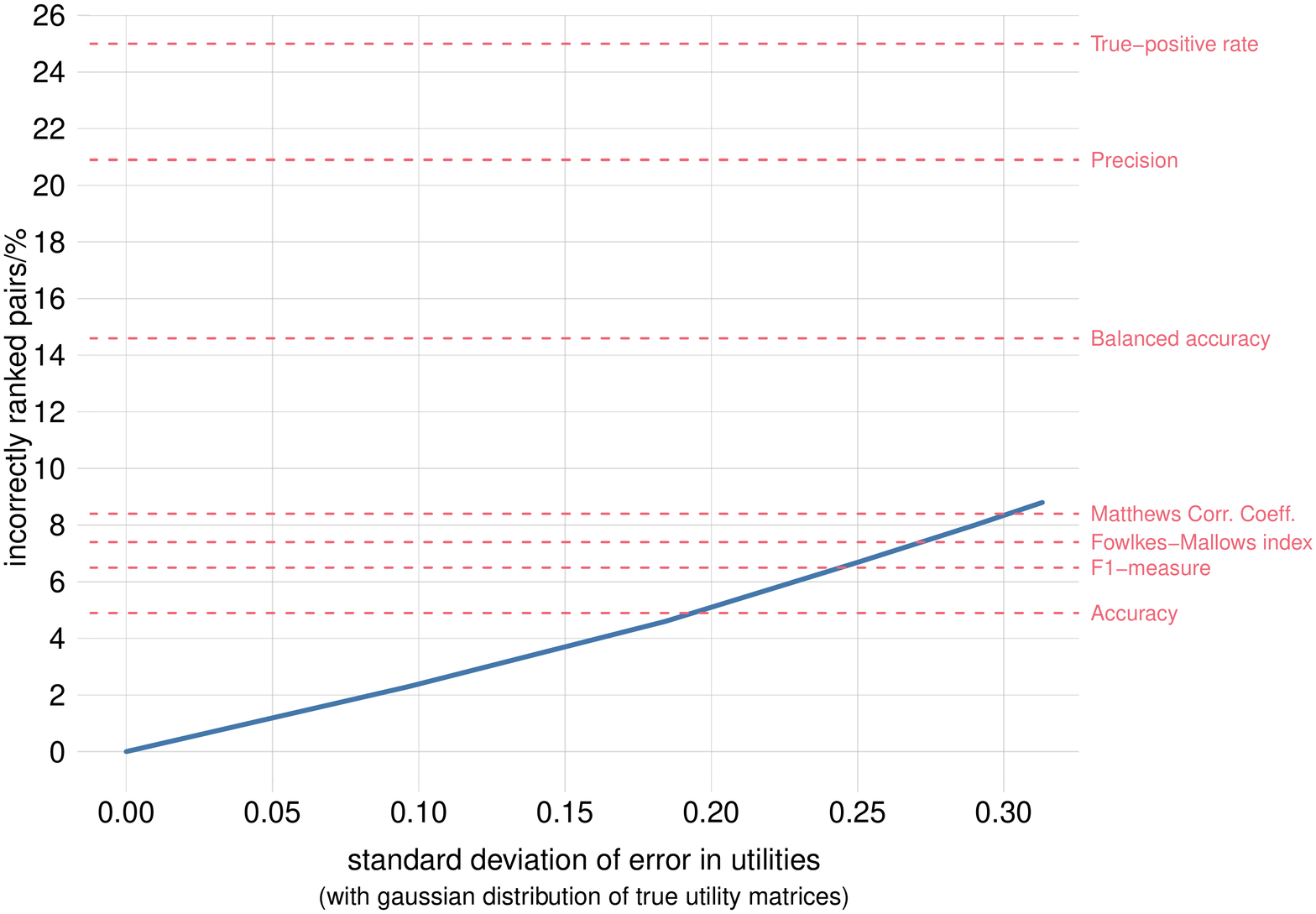}\\\hfill
  \caption{Dependence of the proportion of incorrectly ranked pairs of confusion matrices, on the standard deviation of the assessment error on the utilities. Top plot: case with uniform distribution of true utility matrices. Bottom plot: case with gaussian distribution of true utility matrices, as in \fig~\ref{fig:gauss_distr_um}.}
  \label{fig:percentage_vs_error}
\end{figure}
Each point represents a pair of confusion matrices (step~\ref{item:draw_cm}); its coordinates are the true utility yield and either the score given by a metric or (last plot) the yield according to the incorrect utility matrix. The red or yellow triangular points in the II and IV quadrants (discordant signs) are incorrectly ranked pairs. The percentages of incorrect rankings are calculated from $10^{6}$ samples, giving slightly more than one decimal significant digit; fewer samples are shown in the plots.

The plots are displayed in order (left-right, top-bottom) of decreasing percentages of incorrect rankings. The accuracy metric proves to be the best among the ones considered, leading to $8.7\%$ incorrect pairwise rankings. But we see that a utility matrix affected by gaussian errors with $0.1$ standard deviation is even better, yielding $4\%$ incorrect pairwise rankings.

The dependence of the fraction of incorrect rankings on the standard deviation of the error affecting the utilities is shown in the plots of \fig~\ref{fig:percentage_vs_error}, for the case of uniform distribution (top plot) and gaussian distribution (bottom plot) of true utility matrices. It is approximately linear. The plots also report the fractions of incorrect rankings for the other metrics. We see that evaluations based on a utility matrix affected by errors with standard deviation up to $0.15$ or even $0.25$ are still more reliable than evaluations based on the other reported metrics. This is a remarkable fact, considering that errors with such standard deviations are quite large, as was shown in \fig~\ref{fig:error_distr_um}.

A utility error with standard deviations around $0.25$ covers the whole space of utility matrices almost uniformly (cf \fig~\ref{fig:error_distr_um}). Such a large error means that we are almost completely uncertain about the utilities to start with. It therefore makes sense that the accuracy, equivalent to using the identity utility matrix, becomes a more reliable metric when this error level is reached: as we saw in \sect~\ref{sec:unknown_utilities}, the identity utility matrix is the natural one to use in a state of complete uncertainty about the utilities. This result is just another example of the internal consistency of Decision Theory.

\section{What about the area under the curve of the receiver operating characteristic?}
\label{sec:auc}

Another very common metric for evaluating binary classifiers is the Area Under the Curve of the Receiver Operating Characteristic, or \enquote{area under the curve} for short. This metric can only be used for particular classifying algorithms, and its meaning is different from that of the metrics reviewed so far. For these reasons, we leave a full discussion of it to future works and only offer a couple of remarks here.

The area under the curve can only be computed for classifiers that output a continuous variable rather than a class. A threshold for this variable determines whether its value predicts one class or the other. Different choices of threshold lead to different pairs of false-positive rate $f$ (which is $1-{}$true-negative rate) and true-positive rate $t$ in a given test set. These pairs can be plotted as a curve $f \mapsto t(f)$ on a graph with corresponding axes. Given the proportion of classes in the test set, every point on the curve corresponds to a possible confusion matrix $C_{ij}(f)$ that the classifier can produce depending on the threshold chosen. The area subtended by the curve is a weighted average of true-positive rates with a peculiar choice of weights; the weights are uniform as a function of the false-positive rate, but generally not uniform as a function of the threshold, for example. The meaning and proper use of the receiver operating characteristic are discussed in a classic by \cites{metz1978}, see especially p.~290.

\medskip

From the standpoint of decision theory, two remarks can be made\autocites[similar points are made by][]{bakeretal2001,loboetal2008}. First, according to the principle of maximum expected utility, \sect~\ref{sec:dt_utilities}, we should choose a threshold and corresponding false-positive rate $f^{*}$ such as to maximize the utility yield, given by \eqn~\eqref{eq:final_utility}:
\begin{equation}
  \label{eq:threshold_auc_maxutility}
  \text{\small choose}\quad
  f^{*} = \argmax_{f}\set[\bigg]{\sum_{i,j=0}^{1} U_{ij}\ C_{ij}(f)} \ .
\end{equation}
Any other values of $f$ and of the threshold are irrelevant. Averages over $f$ values are therefore irrelevant as well. Second, suppose our goal is to evaluate the average performance over several possible classification problems. In that case, the quantities to be averaged are the utility matrices of those classification problems, as discussed in \sect~\ref{sec:unknown_utilities}, yielding a unique expected utility matrix. Once this is computed, we go back to a single choice of $f$ according to our first remark.

Owing to these issues, the area under the curve suffers from the same problems as the non-compliant metrics discussed in \sect~\ref{sec:common_metrics}: in every classification problem, it always leads to cases of incorrect evaluation.

A correct use of the receiver-operating-characteristic curve $t(f)$ can be made, however. It is explained in \cite[section \emph{Cost/Benefit Analysis} p.~295]{metz1978}, and in \cite[\sect~5.7.4]{soxetal1988_r2013} (curiously Sox \etal\ also mention the generally erroneous criterion of the area under the curve).

Denote the proportion of class~$0$ (positive) in the test set by $B$. The confusion matrix as a function of $f$ is then
\begin{equation}
  \label{eq:CM_fromrates}
  \umatrix{
    C_{00}(f) }{ C_{01}(f)}{ C_{10}(f) }{ C_{11}(f)
  }
  =
  \begin{bmatrix}
    B\ t(f) & (1-B)\ f \\ B\ [1-t(f)] &(1-B)\ (1-f)
  \end{bmatrix} \ .
\end{equation}
The sum in formula~\eqref{eq:threshold_auc_maxutility} above can then be explicitly written, rearranging some terms,
\begin{multline}
  \label{eq:sum_UC}
  \sum_{i,j=0}^{1} U_{ij}\ C_{ij}(f) \equiv
      (U_{00} - U_{10})\  B\ t(f) -
    (U_{11} - U_{01})\ (1-B)\ f
    +{}\\[-2\jot]
    U_{10}\ B\ + U_{11}\ (1-B) \ .
\end{multline}
The principle of maximum expected utility~\eqref{eq:threshold_auc_maxutility} is  then equivalent to the following condition, obtained using the explicit sum above but dropping the constant term on the second line for simplicity:
\begin{equation}
  \label{eq:threshold_auc_maxutility_rates}
  \text{\small choose}\quad
  f^{*} = \argmax_{f}\set[\big]{
    (U_{00} - U_{10})\  B\ t(f) -
    (U_{11} - U_{01})\ (1-B)\ f
} \ .
  \end{equation}
The function in braces is monotonically increasing because $t(f)$ is (we assume, as always, that the utility of correct classification of a class is higher than that of misclassification, so $U_{00}-U_{10} \ge 0$ and $U_{11}-U_{01} \ge 0$). Its maximum can thus be found by setting its derivative to zero:
\begin{equation}
  \label{eq:threshold_auc_maxutility_derivat}
  \text{\small choose } f^{*} \text{\small\ such that}\quad
t'(f^{*}) = \frac{(U_{11} - U_{01})\ (1-B)}{(U_{00} - U_{10})\  B} \ .
\end{equation}
\begin{figure}
  \centering
    \includegraphics[width=0.66\linewidth]{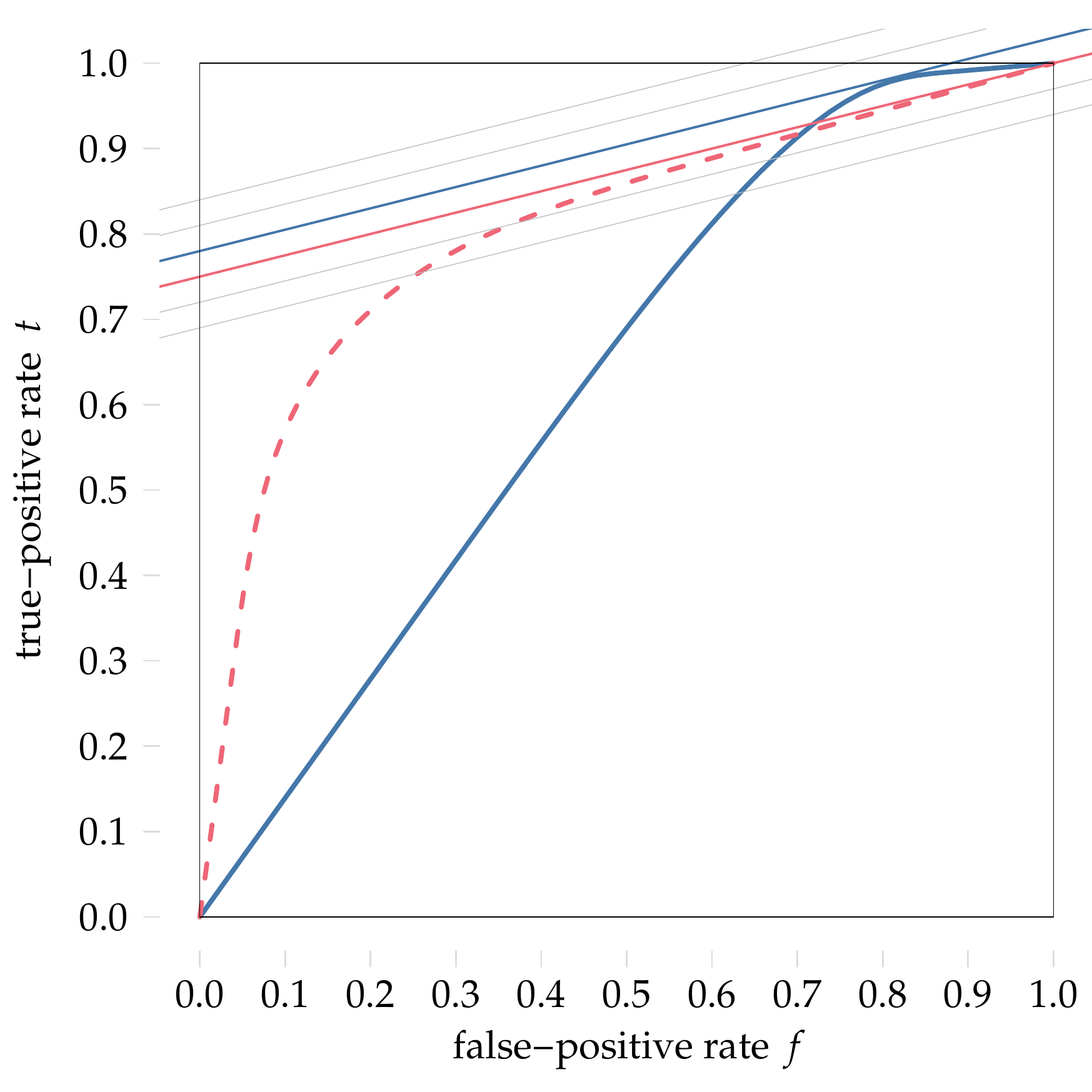}\\
    \caption{Receiver-operating-characteristic curves of two classifiers. The {\color{red}red dashed curve} clearly subtends a larger area than the {\color{blue}blue solid curve}. Yet the classifier with the latter curve yields a higher utility, because it touches the family of parallel lines, \eqn~\eqref{eq:tangent_line}, at a higher point. This example arises for a utility matrix equal to $\sumatrix{ 4}{0}{0}{1 }$ and a test set with $B=0.5$ (balanced), or for a utility matrix equal to $\sumatrix{ 1}{0}{0}{1 }$ and a test set with $B=0.8$.}
  \label{fig:auc_example}
\end{figure}
If we have several classifiers, each with its own curve $t(f)$, then \emph{the best is the one tangent to the line
\begin{equation}
  \label{eq:tangent_line}
  t = \frac{(U_{11} - U_{01})\ (1-B)}{(U_{00} - U_{10})\  B} f + \text{\normalfont\small const.}
\end{equation}
that has the highest intercept.}

From this criterion it can be seen geometrically that if a classifier has its curve $t(f)$ \emph{completely above} the curve of another classifier, then it must have a higher utility yield. But nothing, in general, can be said if the curves of the two classifiers cross. It is the tangent of a receiver-operating-characteristic curve that matters, not its subtended area. Figure~\ref{fig:auc_example} shows an example of this.

%

\section{Summary and discussion}
\label{sec:discussion}

The evaluation and ranking of classification algorithms is a critical stage in their development and deployment. Without such evaluation we cannot even say whether an algorithm is better than another, or whether a set of parameter values for a specific algorithm is better than another set.

And yet, at present, we have not an evaluation theory but only an evaluation folklore: different procedures, proposed only out of intuition and of analysis of special cases, with fuzzy criteria to decide which should be used, and without rigorous theoretical foundations that should guarantee uniqueness, universality properties, and absence of biases. We believe that some of the surprising failures of machine learning \emph{in actual applications}\autocites[see \eg][]{varoquauxetal2022} come not only from biases in the choice of test datasets and other similar biases, but also from the use of wrong evaluation metrics in the development stage.

In the present work, we have argued that theoretical foundations for the evaluation process are available in \emph{Decision Theory}. Its main notions and principle -- utilities and their maximization -- are very intuitive, as shown (we hope) by the introductory story.

These are the main results of the application of decision theory to the evaluation of classifiers:
\begin{itemize}[wide]
\item The evaluation metric must depend on the specific classification problem.
\item Such metric is completely defined by $n^{2}$ parameters, called utilities, collected in a utility matrix; $n$ is the number of classes. Two parameters are arbitrary and represent a zero and measurement unit of the utility scale. In the binary-classification case, this means that we have a two-dimensional set of possible metrics.
\item The score of a classifier on a test set is simply given by its utility yield: the grand sum of the products of the elements of the utility matrix and the confusion matrix of the classifier. It is a simple linear expression in the confusion-matrix elements.
\item A utility matrix, obtained from an average, is also used when we are uncertain about the utilities underlying a classification problem or when we want to consider the average performance over several classification problems.
\item Some popular metrics such as precision, balanced accuracy, Matthews correlation coefficient, Fowlkes-Mallows index, $F_{1}$-measure, and area under the receiver-operating-characteristic curve do not comply with decision theory. As a consequence, they always lead to some erroneous comparative evaluations of classifiers in every classification problem, even when all utilities and frequencies are correctly assessed; are they are likely affected by cognitive biases.
\item Using a utility matrix with incorrectly assessed utilities still leads, on average, to fewer wrong comparative evaluations than using other popular metrics.
\end{itemize}

We believe that the decision-theoretic evaluation of classifiers also has remarkable advantages:

First, it translates the fuzzy problem \enquote*{which of the numerous scores should I rely on?} into a more structured, thus easier to confront, one: to assess, at least semi-quantitatively, how many times more valuable, desirable, or useful is the correct classification of a class than its incorrect classification, than the correct classification of another class, and so on. Such utilities usually have a more immediate, problem-dependent interpretation than other metrics.

Second, it leads to a mathematically simple, computationally convenient metric: a linear combination of confusion-matrix elements -- no need for non-linear functions or integration of curves.

Third, the principles of the underlying theory guide us if we have to face new peculiar problems. Imagine, for instance, a classification problem where we cannot say, in general, whether true positives are more important than true negatives and so on, because such valuation can \emph{vary from one tested item to another}. Decision theory, in this case, requires an item-wise assessment of utilities, and still provides an item-wise score, which can be accumulated across items to obtain a total evaluation score for the performance of candidate classifiers.

\medskip

The theory, remarks, and results of the present work generalize beyond classification: to regression and more complex classification-like problems such as image segmentation, with important applications in medicine \autocites{lundervoldetal2019}. It would be interesting to examine whether popular metrics in the latter field, such as Dice score \autocites{dice1945,fleiss1975,zijdenbosetal1994} and Hausdorff distance \autocites{altetal2000}, comply with decision-theoretic principles, and which alternatives could be used otherwise.

In a companion work \autocites{dyrlandetal2022b} we apply the general ideas presented here to improve the performance of \ml\ classifiers.




\begin{contributions}
  All authors have contributed equally to the present work.
\end{contributions}
\begin{acknowledgements}
  KD and ASL acknowledge support from the Trond Mohn Research Foundation, grant number BFS2018TMT07, and PGLPM from The Research Council of Norway, grant number 294594.

  KD would like to thank family for endless support; partner Synne for constant love, support, and encouragement; and the developers and maintainers of Python, FastAi, PyTorch, scikit-learn, NumPy and RDKit for free open source software and for making the experiments possible. 

  PGLPM thanks Iv{\'an} Davidovich for useful comments on previous drafts of this work; Maja, Mari, Miri, Emma for continuous encouragement and affection; Buster Keaton and Saitama for filling life with awe and inspiration; and the developers and maintainers of \LaTeX, Emacs, AUC\TeX, Open Science Framework, R, Python, Inkscape, LibreOffice, Sci-Hub for making a free and impartial scientific exchange possible.
\end{acknowledgements}


\renewcommand*{\finalnamedelim}{\addcomma\space}
\defbibnote{prenote}{{\footnotesize (\enquote{de $X$} is listed under D,
    \enquote{van $X$} under V, and so on, regardless of national
    conventions.)\par}}

\printbibliography[prenote=prenote
]

\end{document}